\documentclass[10pt,journal,compsoc]{IEEEtran}
\ifCLASSOPTIONcompsoc
  \usepackage[nocompress]{cite}
\else
  \usepackage{cite}
\fi
\ifCLASSINFOpdf
\else
\fi

\usepackage{amsmath}
\usepackage{amssymb}
\usepackage{graphicx}
\usepackage{multirow}
\usepackage{pifont}
\usepackage[colorlinks]{hyperref}

\newcommand{\cmark}{\ding{51}}
\newcommand{\xmark}{\ding{55}}

\begin{document}
%
\title{MutualNet: Adaptive ConvNet via Mutual Learning from Different Model Configurations}
%
%
%
%

\author{Taojiannan Yang,
        Sijie Zhu,
        Matias Mendieta,
        Pu Wang,
        Ravikumar Balakrishnan,
        Minwoo Lee,
        Tao Han,
        Mubarak~Shah,~\IEEEmembership{Fellow,~IEEE,}
        and~Chen~Chen,~\IEEEmembership{Member,~IEEE}
\IEEEcompsocitemizethanks{\IEEEcompsocthanksitem T. Yang, S. Zhu, M. Mendieta, C. Chen, and M. Shah are with the Center for Research in Computer Vision, University of Central Florida, Orlando, FL 32816. E-mail: taoyang1122@knights.ucf.edu
\IEEEcompsocthanksitem P. Wang and M. Lee are with the Department
of Computer Science, University of North Carolina at Charlotte, Charlotte,
NC 28223.
\IEEEcompsocthanksitem R. Balakrishnan is with Intel Labs.
\IEEEcompsocthanksitem T. Han is with the Department of Electrical and Computer Engineering, New Jersey Institute of Technology, Newark, NJ 07102.
\IEEEcompsocthanksitem This work is partially supported by the National Science Foundation (NSF) under Grant No. 1910844 and NSF/Intel Partnership on MLWiNS under Grant No. 2003198.  \protect\\
}
}

%
%

\markboth{Journal of \LaTeX\ Class Files,~Vol.~14, No.~8, August~2015}%
{Shell \MakeLowercase{\textit{et al.}}: Bare Demo of IEEEtran.cls for Computer Society Journals}
%



\IEEEtitleabstractindextext{%
\begin{abstract}
Most existing deep neural networks are static, which means they can only perform inference at a fixed complexity. But the resource budget can vary substantially across different devices. Even on a single device, the affordable budget can change with different scenarios, and repeatedly training networks for each required budget would be incredibly expensive. Therefore, in this work, we propose a general method called MutualNet to train a single network that can run at a diverse set of resource constraints. Our method trains a cohort of model configurations with various network widths and input resolutions. This mutual learning scheme not only allows the model to run at different width-resolution configurations but also transfers the unique knowledge among these configurations, helping the model to learn stronger representations overall. MutualNet is a general training methodology that can be applied to various network structures (e.g., 2D networks: MobileNets, ResNet, 3D networks: SlowFast, X3D) and various tasks (e.g., image classification, object detection, segmentation, and action recognition), and is demonstrated to achieve consistent improvements on a variety of datasets. Since we only train the model once, it also greatly reduces the training cost compared to independently training several models.  Surprisingly, MutualNet can also be used to significantly boost the performance of a single network, if dynamic resource constraints are not a concern. In summary,  MutualNet is a unified method for both static and adaptive, 2D and 3D networks. Code and pre-trained models are available at \url{https://github.com/taoyang1122/MutualNet}.
\end{abstract}

}

\maketitle

\IEEEdisplaynontitleabstractindextext

%
\IEEEpeerreviewmaketitle

\IEEEraisesectionheading{\section{Introduction}\label{sec:introduction}}

\IEEEPARstart{D}{eep} neural networks have triumphed over various perception tasks including image classification \cite{resnet, alexnet, mobilenet}, object detection \cite{liu2016ssd, fasterrcnn}, semantic segmentation \cite{fcn, chen2017deeplab} and so on. However, most existing deep neural networks are static, which means they can only run at a specific resource constraint. For example, MobileNet \cite{mobilenet} has 4.2M model parameters and 569M FLOPs. After training, the model can only do inference at this specific complexity. If we want to reduce the model complexity, we have to re-train a smaller model; otherwise, the performance will drop substantially. As shown in Table \ref{tab:reducewidth}, a regular MobileNet has 70.6\% Top-1 accuracy on ImageNet. However, if we only use half of its channels ($width=0.5\times$) to do inference, the performance  drops to 0.4\%. This is almost the same accuracy as a simple random guess. But if we re-train the MobileNet-$0.5\times$ from scratch, it can achieve 63.3\% Top-1 accuracy \cite{mobilenet}. A similar trend is observed if the network width is unchanged ($width=1.0\times$) while the input image resolution is reduced. Although the performance does not drop as dramatically as by reducing the network width, the gap between the smaller resolution and full resolution is still quite large. These results indicate that regular deep neural networks can not generalize well to other network widths and image resolutions, and restrains their effectiveness to a specific resource budget.

\begin{table}[!t]
    \renewcommand{\arraystretch}{1.2}
    \caption{Reducing MobileNet complexity by width or resolution at runtime. The network fails to achieve a good performance without re-training.}
    \label{tab:reducewidth}
    \centering
    \resizebox{\linewidth}{!}{
    \begin{tabular}{l|cc|cc|cc}
    \hline
    \multicolumn{1}{l|}{Width} & \multicolumn{2}{c|}{$1.0\times$} & \multicolumn{2}{c|}{$0.75\times$} & \multicolumn{2}{c}{$0.5\times$} \\
    \multicolumn{1}{l|}{re-train} & \cmark & \xmark & \cmark & \xmark & \cmark & \xmark \\
    \hline
    Acc (\%) & 70.6 & 70.6 & 68.4 & 14.2 & 63.3 & 0.4 \\
    \hline
    \hline
    \multicolumn{1}{l|}{Resolution} & \multicolumn{2}{c|}{$224 \times 224$} & \multicolumn{2}{c|}{$160 \times 160$} & \multicolumn{2}{c}{$128 \times 128$} \\
    \multicolumn{1}{l|}{re-train} & \cmark & \xmark & \cmark & \xmark & \cmark & \xmark \\
    \hline
    Acc (\%) & 70.6 & 70.6 & 67.2 & 65.0 & 64.4 & 57.7 \\  
    \hline
    \end{tabular}
    }
\end{table}

In real-world applications, however, the computing capacity of different devices can vary significantly. A model may be small for a high-end GPU but too heavy to run on mobile devices. A common practice is to adopt a global width multiplier \cite{mobilenet, sandler2018mobilenetv2, zhang2018shufflenet} to adjust the model size, but still, models of different scales need to be re-trained for many devices. Besides, even on the same device, the resource budgets can change. For example, the battery condition of mobile devices imposes constraints on the computational budget of many operations.
Similarly, a task may have specific priorities at any given time, requiring a dynamic computational budget throughout its deployment phases.
To meet various resource constraints, one needs to deploy several different scaled networks on the device and switch among them. However, this naive solution is highly inefficient and not scalable due to two main reasons. First, deploying several models will have a much higher memory footprint than a single model. Second, if a model has to cover a new resource constraint, a new model has to be re-trained and deployed on the device. Previous works \cite{slimnet, usnet} propose to train a network that can run at different network widths. However, they ignore the input dimension, thus failing to achieve a practical accuracy-efficiency trade-off, and their method is only evaluated on image classification. In this paper, we propose a unified framework to mutually learn a single  network, which can run at different network widths (network scale) and input resolutions (input scale). The effectiveness of our method is thoroughly evaluated on different 2D and 3D (spatio-temporal) network structures on different tasks.

We first start with 2D deep neural networks where the inputs are images. The inference complexity of a 2D network is determined by both the network scale and input scale. In our approach, we consider different width-resolution configurations for different accuracy-efficiency trade-offs. Our goal is to train a network that can generalize well to various width-resolution configurations. Therefore, our training framework only requires minor changes to the regular training process, making it simple yet effective. During the training phase, besides training the full model (e.g., $1.0\times$-$224$ (width-resolution) on ImageNet), we additionally sample several other configurations (sub-networks by the network width and images of different sizes) and train them jointly with the full model. 
Note that the sub-networks share the weights of the full-network, so this does not introduce extra model parameters. This mutual training framework not only enables the model to do inference at different width-resolution configurations but also facilitates knowledge transfer among the configurations and further enhances their overall performance. In Section \ref{sec:2d-knowledge}, we demonstrate that distinct sub-networks may make their predictions based on different semantic regions. Due to the weight sharing strategy, sub-networks can naturally share their knowledge with each other, which helps all the sub-networks to learn more diversified representations. 

Apart from 2D networks, 3D (spatio-temporal) networks have achieved excellent performance on video tasks, but they generally require more computation (tens or even hundreds of GFLOPs
per video clip) and training time than 2D networks on large-scale datasets. 
This makes training and deploying several 3D networks even less practical than the 2D case.
Therefore, we further extend our method to 3D networks where the inputs are videos instead of images. Most 3D networks \cite{c3d, i3d, 3dresnet, x3d} are driven by expanding 2D convolutional layers in image architectures \cite{resnet, alexnet, googlenet} into 3D convolutions. For these direct extensions, our method can be easily applied by jointly sampling the spatial and temporal dimensions. Recently, SlowFast \cite{slowfast} proposes a two-branch structure to deal with spatial and temporal dimensions asymmetrically. The Slow branch captures spatial semantics while the Fast branch captures fine temporal information. Following this idea, we conduct an asymmetrical sampling where we only sample sub-networks on the Slow branch, while keeping the Fast branch unchanged to provide complementary information for the adaptive Slow branch.

MutualNet is simple but surprisingly effective. On image classification tasks, our method significantly outperforms previous adaptive networks \cite{slimnet, usnet} on various structures. Surprisingly, it also outperforms individually-trained models at different configurations. Thanks to its simplicity, MutualNet can be easily applied to other tasks including object detection, instance segmentation and action recognition. To the best of our knowledge, we are the first ones to demonstrate adaptive networks on these tasks, which also  achieve consistent improvements over individually-trained networks. We further demonstrate that MutualNet is promising methodology to serve as a general training strategy to boost the performance of a single network. It achieves comparable or better performance than state-of-the-art training techniques, e.g., data augmentation \cite{zhang2017mixup, cubuk2019autoaugment}, regularization \cite{cutout, yamada2019shakedrop}, SENet \cite{senet}.

This work is an extension of the conference version \cite{yang2020mutualnet}. In this paper, we make the following important extensions:
\begin{itemize}
    \item We present more results on ImageNet image classification with different network architectures (e.g., MobileNetv1, MobileNetv2, ResNet-50). Our method consistently outperforms previous adaptive networks \cite{slimnet, usnet} and individually-trained networks.
    \item We extend the method to 3D (spatio-temporal) networks for action recognition. We design two sub-network sampling strategies for one-branch and two-branch structures.
    \item We conduct extensive experiments to evaluate the effectiveness of MutualNet on 3D networks. Our method outperforms state-of-the-art networks (e.g., SlowFast, X3D) on Kinetics-400 at different resource budgets. The effectiveness of the learned representation is also validated via cross-dataset transfer (Charades dataset \cite{charade}) and cross-task transfer (action detection on AVA dataset \cite{gu2018ava}).
    \item We present an analysis of the theoretical complexity of our method. We also report the wall-clock timings in practice. Our method requires just a fraction of the training cost compared to independently training several models. For inference, MutualNet deploys a single  model rather than several individually-trained models to cope with dynamic budgets.
\end{itemize}

The remainder of the paper is organized as follows. In Section \ref{sec:relatedwork}, we review some related works. In Section \ref{sec:2dmutualnet} and Section \ref{sec:3dmutualnet}, we introduce MutualNet on 2D and 3D networks respectively. In Section \ref{sec:complexity}, we analyze and compare the training complexity of MutualNet and independent networks. Experimental results are compared in Section \ref{sec:experiments}. Finally, in Section \ref{sec:conclude}, we conclude our work.

\section{Related Work}\label{sec:relatedwork}
{\bf Light-weight Networks.} 
There has recently been a flurry of interest in designing light-weight networks. MobileNets~\cite{mobilenet, sandler2018mobilenetv2} factorize the standard $3\times3$ convolution into a $3\times3$ depthwise convolution and a $1\times1$ pointwise convolution which reduces computation cost by several times. ShuffleNets~\cite{zhang2018shufflenet, shufflenetv2} separate the $1\times1$ convolution into group convolutions and shuffle the groups to further improve accuracy-efficiency trade-offs.
ShiftNet~\cite{shiftnet} introduces a zero-flop shift operation to reduce computation cost. AdderNet \cite{chen2020addernet} trades the massive multiplications in deep neural networks for much cheaper additions. GhostNet \cite{han2020ghostnet} leverages cheap linear transformations to generate more ghost feature maps. Most recent works \cite{wu2019fbnet, tan2019mnasnet, mobilenetv3} also apply neural architecture search methods to search efficient network structures. However, none of them consider the varying resource constraints during runtime in real-world applications. \textit{To meet different resource budgets, these methods need to train and deploy several models and switch among them, which is inefficient and not scalable}.

{\bf Adaptive/Dynamic Neural Networks.} There is a growing interest in the area of dynamic and adaptive neural networks. One category of methods \cite{msdnet, resolutionadaptive, glancefocus, arnet} perform dynamic inference conditioned on the input. The core principle of these methods is to utilize less computation for easy samples and reserve more computation for hard ones. MSDNet~\cite{msdnet} proposes a multi-scale and coarse-to-fine densenet framework. It has multiple classifiers and can make early predictions for easy instances. RANet \cite{resolutionadaptive} motivates its design with the idea that low-resolution images are enough for classifying easy samples, while only hard samples need high-resolution input. GFNet \cite{glancefocus} processes a sequence of small patches from the original images and terminates inference once the model is sufficiently confident about its prediction. There are also many recent works \cite{arnet, wu2019adaframe, meng2020adafuse, vared, wu2019liteeval} aiming to reduce spatial and temporal redundancies in videos for action recognition by dynamically processing input frames and fusing feature maps. Inspired by SENet \cite{senet}, a series of works \cite{yang2019condconv, dynamicconv, dyrelu} also propose to learn dynamic attention for different samples. Our method is closer to another category of methods \cite{slimnet, usnet, kim2018nestednet, onceforall, yu2020bignas} where the dynamic routes are determined by the resource budgets. NestedNet~\cite{kim2018nestednet} uses a nested sparse network consisting of multiple levels to meet various resource requirements. SlimmableNet \cite{slimnet, usnet} proposes to train several sub-networks together and perform inference at different network widths. However, it fails to achieve a strong accuracy-efficiency trade-off since it ignores the input dimension. Later works \cite{onceforall, yu2020bignas} further integrate other dimensions (e.g., depth and kernel size) into the training framework, but they do so with neural architecture search (NAS) and thereby require a very complex and expensive training process. Furthermore, their effectiveness is only evaluated on image classification, while our method is thoroughly evaluated on image classification, detection, segmentation and action recognition.

\textbf{Spatio-temporal (3D) Networks.}
The basic idea of video classification architectures stems from 2D image classification models. The works reported in\cite{c3d, i3d, 3dresnet} build 3D networks by extending 2D convolutional filters \cite{alexnet, vgg, googlenet, resnet} to 3D filters along the temporal axis; then the 3D filters are used to learn spatio-temporal representations in a similar way to their 2D counterparts. Later works \cite{s3d, p3d, r2+1d, slowfast} propose to treat the spatial and temporal domains differently. The authors in \cite{s3d} posit that a bottom-heavy structure is better than naive 3D structures in both accuracy and speed. In \cite{p3d, r2+1d}, 3D filters are split into 2D+1D filters, which reduce the heavy computational cost of 3D filters and improve the performance. SlowFast \cite{slowfast} further shows that space and time should not be handled symmetrically and introduces a two-path structure to deal with slow and fast motion separately. Several works \cite{multifiber, groupedspatiotemporal, CSN} also leverage the group convolution and channel-wise separable convolution in 2D networks to reduce computational cost. Recently, \cite{nasvideo1, nasvideo2, nasvideo3} explore neural architecture search (NAS) techniques to automatically learn spatio-temporal network structures. However, 
\emph{all these structures are static. We are the first to achieve adaptive 3D networks which also outperform state-of-the-art independently-trained models \cite{slowfast, x3d}}.

\begin{figure}[t]
\centering
    \includegraphics[width=0.95\linewidth]{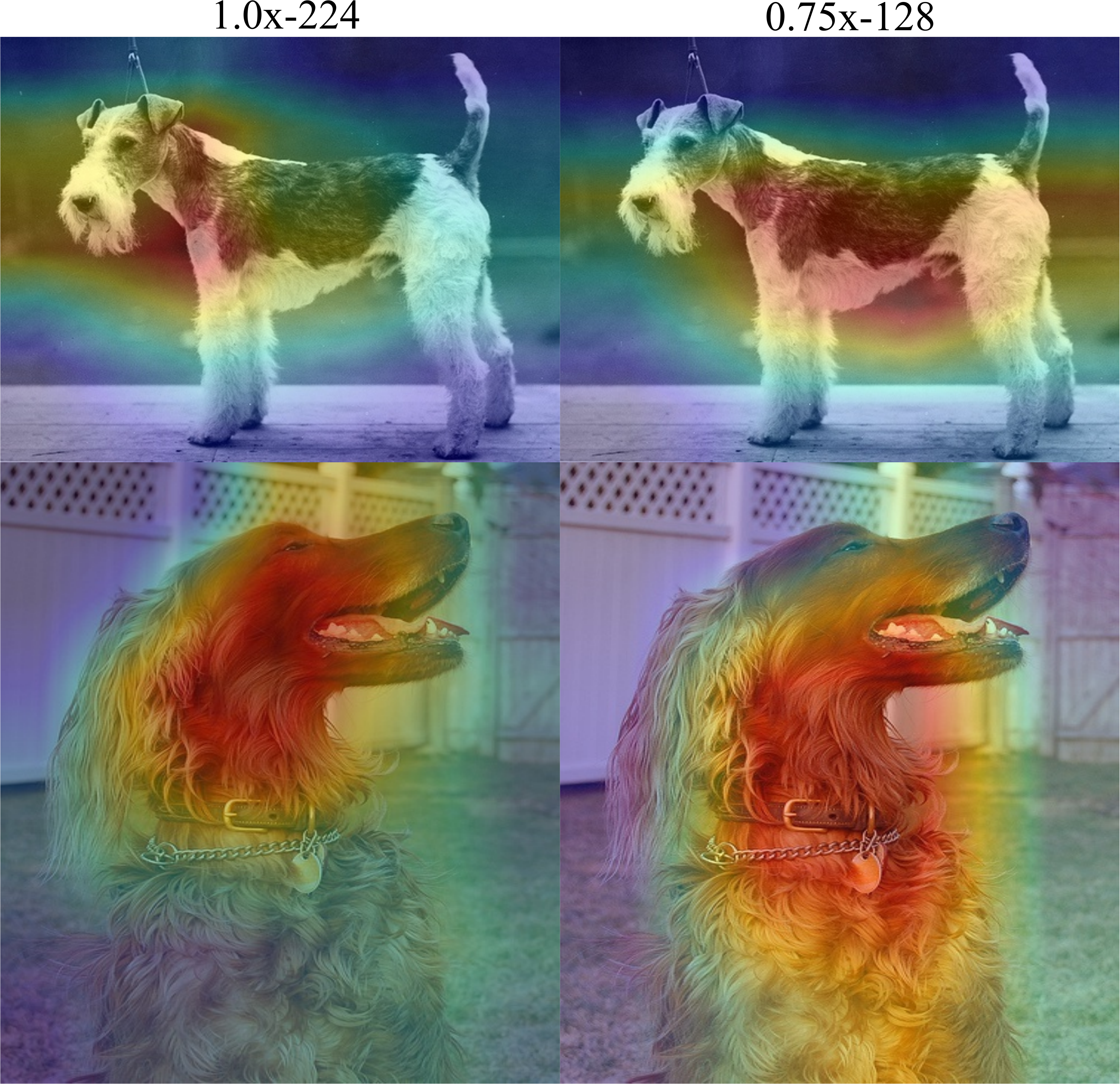}
    \caption{Class activation maps (CAM) of different model configurations (the network is ResNet-50 and is trained on ImageNet). Larger model configuration focuses more on details (e.g., face) while the smaller one focuses more on the contour (e.g., body). }
    \label{fig:2d_cam}
\end{figure}

\begin{figure*}[!t]
    \centering
    \includegraphics[width=0.95\linewidth]{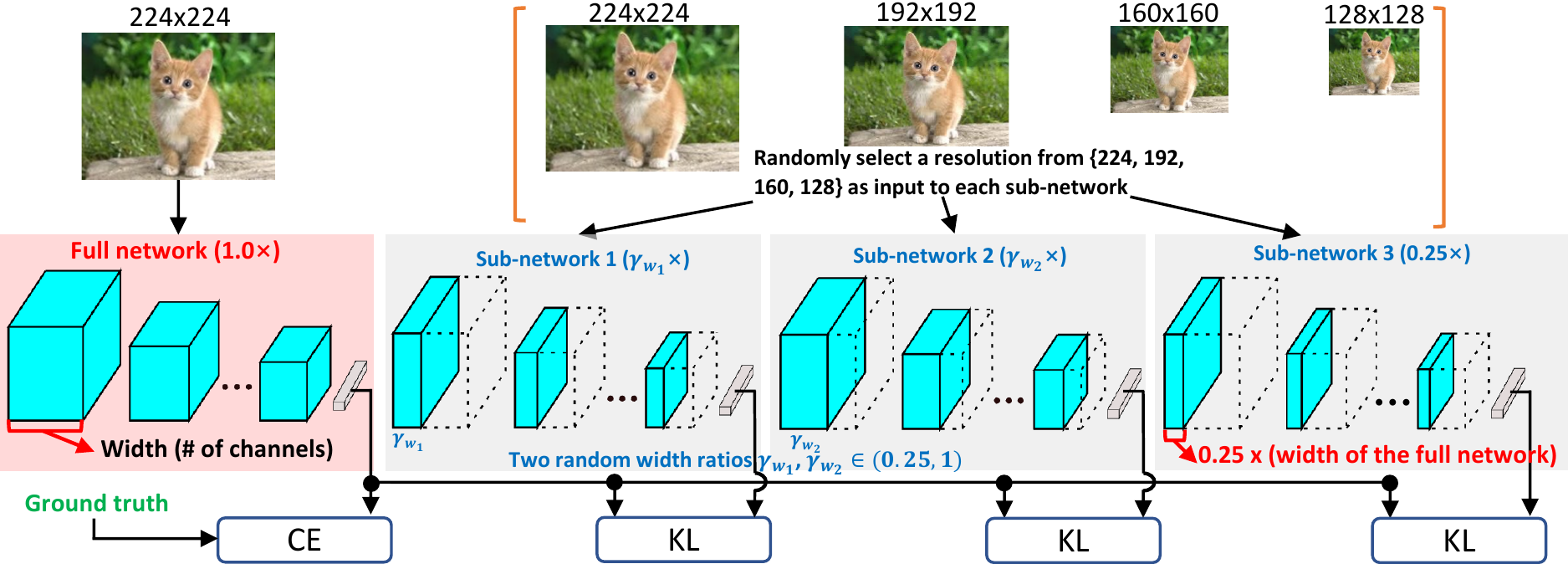}
    \caption{An example to illustrate the training process of MutualNet. The network width range is [0.25$\times$, 1.0$\times$], input resolution is chosen from \{224, 192, 160, 128\}. This can achieve a computation range of [13, 569] MFLOPs on MobileNet v1 backbone. We follow \cite{usnet} to sample 4 networks, i.e., upper-bound full width network ($1.0\times$), lower-bound width network ($0.25\times$), and \textbf{two random width ratios $\gamma_{w_1}, \gamma_{w_2} \in (0.25, 1)$}. For the full-network, we constantly choose 224$\times$224 resolution. For the other three sub-networks, we randomly select its input resolution. The full-network is optimized with the ground-truth label using   Cross Entropy loss (CE). Sub-networks are supervised by the prediction of the full-network using Kullback–Leibler Divergence loss (KL). \textbf{Weights are shared among different networks to facilitate mutual learning}. }
    \label{fig:2d-framework}
\end{figure*}

\textbf{Multi-dimension Trade-off.}
The computational cost and accuracy of a model is determined by both the input size and network size. There is a growing interest \cite{tan2019efficientnet, multidimpruning, x3d} in achieving better accuracy-efficiency trade-offs by balancing different model dimensions (e.g., image resolution, network width and depth). In \cite{multidimpruning}, networks are pruned from multiple dimensions to achieve better accuracy-complexity trade-offs. EfficientNet \cite{tan2019efficientnet} performs a grid-search on different model dimensions and expands the configuration to larger models. X3D \cite{x3d} shares similar ideas with EfficientNet but focuses on spatio-temporal networks. It expands a 2D network along different dimensions to a 3D one. In \cite{tan2019efficientnet, x3d}, the optimal model configuration is searched for a static model and different configurations are trained independently in the process. However, \textit{we aim to learn an adaptive network which can fit various resource budgets.} We train various intrinsic configurations jointly and share knowledge between them, which largely saves the training time and improves the overall performance.

\section{2D MutualNet}\label{sec:2dmutualnet}
Standard 2D models are trained at a fixed width-spatial configuration (e.g., 1.0$\times$-224 on ImageNet). However, the model does not generalize well to other configurations during inference as shown in Table \ref{tab:reducewidth}. In our method, we randomly sample different width-spatial configurations during training, so the model can run effectively at various configurations during inference. Note that the computation cost of a vanilla 2D convolutional layer is given by
\begin{equation}
\label{eq:2dcost}
    K\times K \times C_{i}\times C_{o} \times H \times W.
\end{equation}
Here, $K$ denotes the kernel size, and $C_{i}$ and  $C_{o}$ respectively denote the input and output channels of this layer, while  $H$ and $W$ respectively denote the height and width of the output feature map. For a smaller model configuration, e.g. $0.5\times$-160, the width is reduced by $\gamma_w=0.5$ and the spatial resolution is reduced by $\gamma_s=160/224=0.7$. The computation cost is reduced to
\begin{equation}\label{eq:2dreducecost}
    K\times K \times \gamma_{w}C_{i}\times \gamma_{w}C_{o} \times \gamma_{s}H \times \gamma_{s}W,
\end{equation}
which is $\rho=\gamma_{w}^2\gamma_{s}^2$ times that of the original in Eq. \ref{eq:2dcost}. The dynamic execution range of MutualNet is determined by the range of $\gamma_w$ and $\gamma_s$. The detailed settings of $\gamma_{w}, \gamma_{s}$ and $\rho$ will be discussed in the following sections.


In Section \ref{sec:2d-knowledge}, we first show that distinct model configurations focus on different semantic information in an image. Then, we introduce the training process of our method by a concrete example. Section \ref{sec:2d-analysis} explains the working mechanism of mutual learning from the perspective of model gradients. Section \ref{sec:2d-inference} introduces how to deploy the model and do inference at different resource budgets.

\begin{figure*}[t]
\centering
    \includegraphics[width=0.95\linewidth]{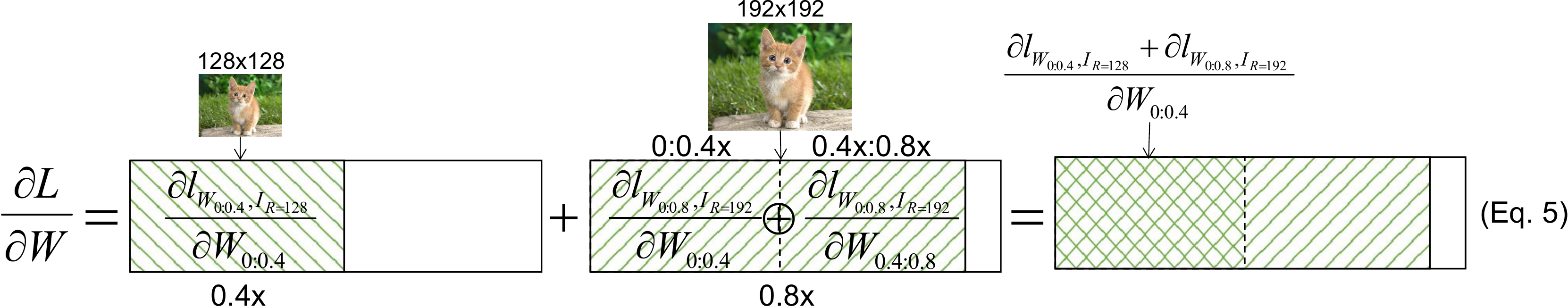}
    \caption{An illustration of the mutual learning scheme. It allows the sub-network to learn multi-scale representations, in terms of both network width and input image resolution.}
    \label{mutual}
\end{figure*}

\subsection{Knowledge in different 2D model configurations}
\label{sec:2d-knowledge}
We want to randomly sample different model configurations in each training iteration to allow them to learn from each other. However, is there any unique knowledge in different model configurations that is beneficial for transferring to others? The answer is yes. Fig. \ref{fig:2d_cam} shows the classification activation maps (CAM) \cite{cam} of two model configurations. The models are trained independently at the corresponding configuration. We can see that these two models focus on different semantic regions of the same object. The larger model configuration (i.e., $1.0\times$-224) tends to focus on fine details (e.g., face of the dog) while the smaller configuration learns the global structures (e.g., the whole body). 
This can be partially attributed to the downsampling of the input resolution, where some fine-grained information is lost but the object contour is enhanced. To further demonstrate that different model configurations have varied attention, we leverage their attention maps to conduct weakly-supervised object localization using CAM \cite{cam} on the ImageNet validation set and compare their localization accuracy on large and small objects. We define small objects as those with a ground truth bounding box smaller than 20\% of the image size, and large objects as those with a ground truth bounding box larger than 50\% of the image size. A prediction is considered correct if its IoU with the ground truth bounding box is larger than 0.5. The results are shown in Table \ref{tab:weaklydet}. We can see that 1.0$\times$-224 achieves the highest accuracy on all objects and large objects, while 1.0$\times$-160 performs better on small objects. Also, 0.75$\times$-128 has lower performance on small objects but higher performance on large objects compared to 1.0$\times$-160. The results show that different configurations focus on different semantic regions. Inspired by this observation, we leverage large configurations to supervise small ones during training. This further enhances the knowledge transfer among different configurations and helps the model to learn more diversified representations.

\begin{table}[t]
\begin{center}
\caption{Weakly-supervised localization accuracy of different model configurations on ImageNet validation set using CAM \cite{cam}. The backbone network is ResNet-50.}
\label{tab:weaklydet}
\begin{tabular}{c|c|c|c}
    \hline
     Model Config & All & Large & Small  \\
     \hline
     1.0$\times$-224 & 37.9\% & 48.3\% & 8.7\%  \\
     1.0$\times$-160 & 25.2\% & 28.3\% & 12.9\% \\
     0.75$\times$-128 & 24.8\% & 31.2\% & 8.3\% \\
     \hline
\end{tabular}
\end{center}
\end{table}

\subsection{2D Model training}
\label{sec:2d-train}
We present an example to illustrate our training process in Fig. \ref{fig:2d-framework}. We set the adaptive width range as [$0.25\times$, $1.0\times$], and the adaptive resolutions as \{224, 192, 160, 128\}. Note that one can adjust these settings according to the required dynamic resource budgets. The depth and resolution can even be larger than the default setting ($1.0 \times$-224) to scale up the model as we show in Section \ref{sec:experiments}. As shown in Fig. \ref{fig:2d-framework}, we first follow \cite{usnet} to sample four sub-networks, i.e., the smallest ($0.25\times$), the largest ($1.0\times$) and \textit{two random width ratios} $\gamma_{w_1}, \gamma_{w_2} \in (0.25, 1)$. Then, unlike traditional ImageNet training with $224\times 224$ input, we resize the input images to resolutions randomly chosen from \{224, 196, 160, 128\} and feed them into different sub-networks. We denote the weights of a sub-network as $W_{0:w}$, where $w \in (0, 1]$ is the width of the sub-network and $0:w$ means the sub-network adopts the first $w \times 100\%$ weights of each layer of the full network. $I_{R=r}$ represents a $r\times r$ input image. Then $N(W_{0:w},I_{R=r})$ represents the output of a sub-network with width $w$ and input resolution $r\times r$. For the largest sub-network (i.e., the full-network in Fig.~\ref{fig:2d-framework}), we always train it with the highest resolution ($224\times 224$) and ground truth label $y$. The loss for the full network is 
\begin{equation}
    loss_{full} = CrossEntropy(N(W_{0:1},I_{R=224}),\, y).
\end{equation}
For the other sub-networks, we randomly pick an input resolution from \{224, 196, 160, 128\} and supervise it with the output of the full-network. As demonstrated in Section \ref{sec:2d-knowledge}, this can transfer the unique knowledge in the full configuration to other configurations and benefit the overall performance. The loss for the $i$-th sub-network is
\begin{equation}
    loss_{sub_i} = KLDiv(N(W_{0:w_i},I_{R=r_i}),\, N(W_{0:1},I_{R=224})),
\end{equation}
where $KLDiv$ is the Kullback-Leibler divergence which measures the distance between two distributions. The total loss is the summation of the full-network and sub-networks, i.e.,
\begin{equation}
    loss = loss_{full}+\sum_{i=1}^3loss_{sub_i}.
\end{equation}
The reason for training the full-network with the highest resolution is that the highest resolution contains more details. Also, the full-network has the strongest learning ability to capture the  discriminatory information from the image data.

\subsection{Gradient analysis of mutual learning}
\label{sec:2d-analysis}
To better understand why the proposed framework can mutually learn from different widths and resolutions, we perform a gradient analysis of the mutual learning process. For ease of demonstration, we only consider two network widths $0.4\times$ and $0.8\times$, and two resolutions 128 and 192 in this example. As shown in Fig. \ref{mutual}, sub-network $0.4\times$ selects input resolution 128, sub-network $0.8\times$ selects input resolution 192. Then we can define the gradients for sub-network $0.4\times$ and $0.8\times$ as $\frac{\partial l_{W_{0:0.4},I_{R=128}}}{\partial W_{0:0.4}}$ and $\frac{\partial l_{W_{0:0.8},I_{R=192}}}{\partial W_{0:0.8}}$, respectively. Since sub-network $0.8\times$ shares weights with $0.4\times$, we can decompose its gradient as 
\begin{equation}
\small
\begin{aligned}
    \frac{\partial l_{W_{0:0.8},I_{R=192}}}{\partial W_{0:0.8}}=\frac{\partial l_{W_{0:0.8},I_{R=192}}}{\partial W_{0:0.4}} \oplus \frac{\partial l_{W_{0:0.8},I_{R=192}}}{\partial W_{0.4:0.8}},
\end{aligned}
\end{equation}
where $\oplus$ is vector concatenation. Since the gradients of the two sub-networks are accumulated during training, the total gradients are computed as
\begin{equation}
\small
\begin{aligned}
    \frac{\partial L}{\partial W} &=\frac{\partial l_{W_{0:0.4},I_{R=128}}}{\partial W_{0:0.4}} + \frac{\partial l_{W_{0:0.8},I_{R=192}}}{\partial W_{0:0.8}} \\
    &= \frac{\partial l_{W_{0:0.4},I_{R=128}}}{\partial W_{0:0.4}} + \left(\frac{\partial l_{W_{0:0.8},I_{R=192}}}{\partial W_{0:0.4}} \oplus \frac{\partial l_{W_{0:0.8},I_{R=192}}}{\partial W_{0.4:0.8}}\right) \\
    &= \frac{\partial l_{W_{0:0.4},I_{R=128}} + \partial l_{W_{0:0.8},I_{R=192}}}{\partial W_{0:0.4}} \oplus \frac{\partial l_{W_{0:0.8},I_{R=192}}}{\partial W_{0.4:0.8}}
\end{aligned}
\end{equation}

\noindent Therefore, the gradient for sub-network $0.4\times$ is $\frac{\partial l_{W_{0:0.4},I_{R=128}} + \partial l_{W_{0:0.8},I_{R=192}}}{\partial W_{0:0.4}}$, which consists of two parts. The first part is computed by itself (0 : 0.4$\times$) with $128\times128$ input resolution. The second part is computed by a larger sub-network $0.8\times$ (i.e., 0 : $0.4\times$ portion) with $192\times192$ input resolution. Thus the sub-network is able to capture multi-scale representations from different input scales and network scales. Due to the random sampling of network width, every sub-network is able to learn multi-scale representations in our framework. This allows the model to  significantly outperform even independently-trained networks. Note that this is different from multi-scale data augmentation as explained in Section \ref{sec:experiments}.

\subsection{2D Model inference}
\label{sec:2d-inference}
After training, the model is able to run at various width-resolution configurations. To deploy the model, we need to find the best-performed model configuration under each particular resource constraint. For evaluation, after following the training example in Section \ref{sec:2d-train}, we first sample the network widths from $0.25\times$ to $1.0\times$ with a step-size of $0.05\times$. Then we sample the input resolutions from \{224, 192, 160, 128\}. We evaluate all these width-resolution configurations on a validation set which gives us a configuration-accuracy table. Similarly, we can evaluate the computational cost (e.g., FLOPs) of each model configuration which gives us a configuration-complexity table. Note that different model configurations may have the same computational cost (e.g., on MobileNet v1, the computational cost of $0.6\times$-224 and $0.7\times$-192 are both $\sim$210 MFLOPs). So the final step is to find the best-performing model configuration for each resource budget, from which we can get the complexity-configuration query table. For real deployment, we only need to deploy the MutualNet model (which is of  the same size as a regular model) and the query table. Then given a resource constraint, we can look up the query table and inference the model at the corresponding optimal configuration. Note that the feature statistics (mean and variance) are different across different model configurations, so we can not use one set of batch normalization (BN) statistics for all configurations. We follow \cite{usnet} to perform BN statistics calibration for each model configuration. Before evaluation, we forward several batches of data to update the BN statistics for a specific configuration. \textbf{There is no re-training so that the whole process is fast and only needs to be done once}.


\begin{figure*}[!t]
    \centering
    \includegraphics[width=0.97\linewidth]{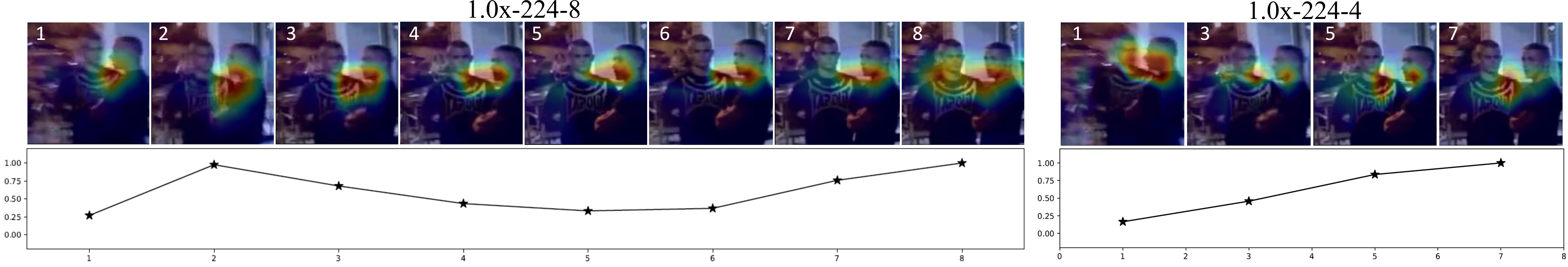}
    \caption{Class activation map along spatial and temporal dimensions of two network configurations. X-axis is the frame index number and y-axis is the normalized activation value. The action is ``headbutting" from the Kinetics-400 dataset.}
    \label{fig:slow_cam}
\end{figure*}
\begin{figure*}[!t]
    \centering
    \includegraphics[width=0.97\linewidth]{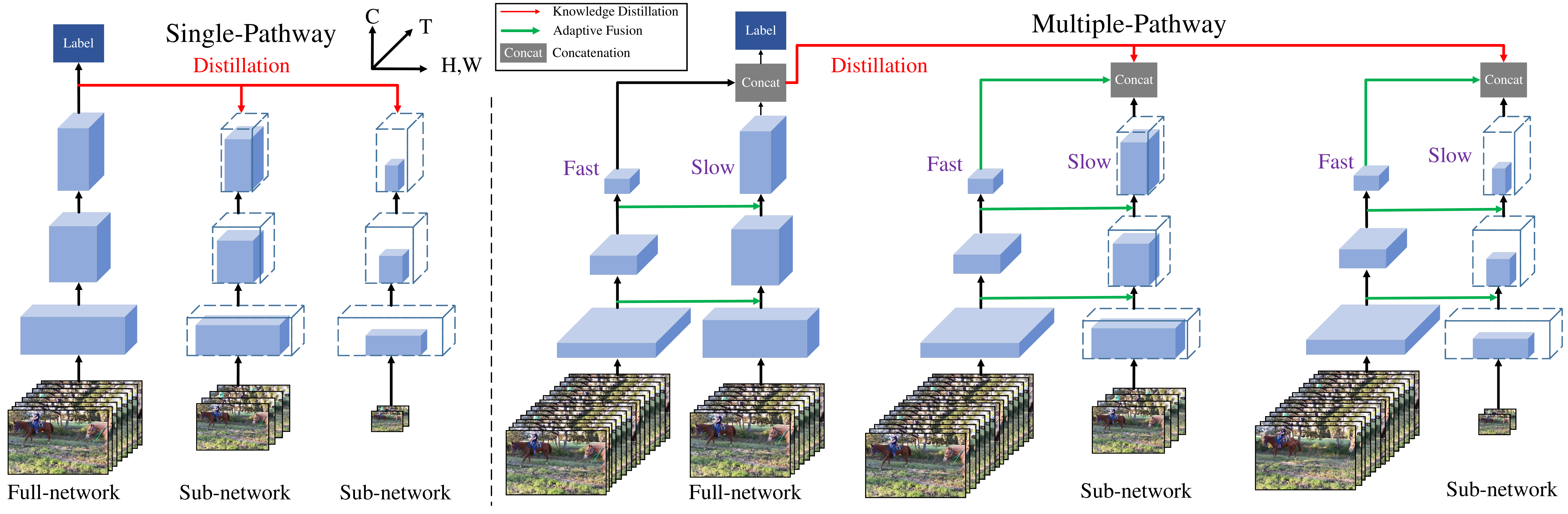}
    \caption{An overview of 3D MutualNet training. Left is for single-pathway structures, which is similar to 2D MutualNet training. Right is for multiple-pathway structures, where only the Slow branch is downsampled. The two branches are fused by the proposed Adaptive Fusion block.}
    \label{fig:3dframework}
\end{figure*}

\section{3D MutualNet}\label{sec:3dmutualnet}
To show the effectiveness of MutualNet as a general training framework, we further extend our method to spatio-temporal (3D) network structures, where the inputs are videos. A popular way to model spatio-temporal information in videos is to extend 2D convolutions to 3D convolutions. The computation cost of a vanilla 3D convolutional layer is given by
\begin{equation}
\label{eq:3dconvcost}
    K \times K \times K \times C_{i}\times C_{o} \times H \times W \times T.
\end{equation}
Here, $K$ denotes the kernel size, and $C_{i}$ and  $C_{o}$ respectively are the input and output channels of this layer. $H$, $W$, $T$ denote the spatial-temporal size of the output feature map. Similar to 2D networks, we sample model configurations by network width, spatial and temporal resolution. Following the notations in Section \ref{sec:2dmutualnet}, we denote a model configuration by width-spatial-temporal. The scaling coefficients $\gamma_w$ and $\gamma_r$ are the same as defined before. $\gamma_t$ is the temporal resolution coefficient where $\gamma_t \in [0, 1]$. The computational cost is reduced to
\begin{equation}\label{eq:3dreducecost}
    K \times K\times K \times \gamma_{w}C_{i}\times \gamma_{w}C_{o} \times \gamma_{s}H \times \gamma_{s}W \times \gamma_{t}T.
\end{equation}
The computational cost is now reduced by $\rho=\gamma_{w}^{2}\gamma_{s}^{2}\gamma_{t}$ times. For single-branch structures \cite{c3d,i3d,x3d}, which are extended from 2D networks, our method can be easily applied by randomly sampling both spatial and temporal dimensions during training. For two-branch structures such as SlowFast \cite{slowfast}, where spatial and temporal dimensions are processed asymmetrically, we conduct asymmetric sampling.

In the following sections, we first show that the temporal dimension can bring additional knowledge to be transferred among different model configurations. Then we explain the training details of one-branch and two-branch structures.

\subsection{Knowledge in different 3D model configurations}
\label{sec:3d-knowledge}
Temporal modeling is essential in 3D networks. Following 2D MutualNet, we first demonstrate that different model configurations will focus on different spatial-temporal semantic regions. Fig. \ref{fig:slow_cam} shows the spatial and temporal distributions of network activation following CAM \cite{cam}. Higher value means more contribution to the final logit value. Although both models generate the prediction as ``headbutting", their decisions are based on different areas of different frames. The input of the left model has $8$ frames, and the 2nd and 8th frames contribute the most to the final prediction. While the right model only has 4 input frames, where those two key frames in the left model are not sampled. So it has to learn other semantic information, forcing a change in both temporal and spatial activation distributions. For example, the activation value of the 5th frame exceeds that of the 3rd frame in the right model, which is the opposite case in the left model. The spatial attention areas are also unique in the two models. In the first frame, the attention is on the shoulder in the left model, but shifts to the face in the right model indicating that a varied set of visual cues is captured.

\subsection{3D Model training}

\subsubsection{One-branch structure}
In one-branch structures, the 3D convolutions are directly extended from 2D convolutions so that we can apply the mutual learning strategy in the same way as 2D networks.
The left half of Fig. \ref{fig:3dframework} shows how mutual training works in single pathway structures. In each training iteration, we sample two sub-networks (by the width factor $\gamma_w$) in addition to the full-network. Sub-networks share the parameters with the full-network in the same way as 2D networks. The full-network is fed with the highest spatial-temporal resolution inputs, while sub-networks are fed with randomly downsampled inputs. Similar to 2D MutualNet training, the full-network is supervised by the ground-truth label while sub-networks are supervised by the full-network to facilitate knowledge transfer. The total loss is the summation of the full-network's loss and sub-networks' losses.

\subsubsection{Two-branch structure}
Recently, SlowFast \cite{slowfast} proposes that the temporal dimension should not be processed symmetrically to the spatial dimension, as slow and fast motions contain different information for identifying an action class.
SlowFast shows that a lightweight fast pathway, which aims to capture fine motion information, is a good complement to the slow pathway, which mainly captures spatial semantics. This inspires us to leverage multiple-pathway trade-offs in 3D MutualNet. The structure is shown in the right half of Fig. \ref{fig:3dframework}. Since the fast pathway is lightweight (about $10\%$ of the overall computation), reducing its spatial-temporal resolution or network width is not beneficial for the overall computation-accuracy trade-off. In two-branch structures, we keep the respective $\gamma_{w},\gamma_{s},\gamma_{t}=1$ for the Fast pathway so that it can provide complementary information for the Slow path with its own $\gamma_{w},\gamma_{s},\gamma_{t}\leq 1$. \emph{Note that this complementary information is not only on temporal resolution but also on spatial resolution.} \\
\noindent\textbf{Adaptive Fusion.}
Given fixed temporal resolutions for two pathways, the fusion is conducted by lateral connections with time-strided convolution in SlowFast \cite{slowfast}.  \emph{However, since all the three dimensions ($\gamma_{w}$, $\gamma_{s}$, $\gamma_{t}$) of the Slow pathway can change during training, directly applying the time-strided convolution does not work for our framework.} Therefore, we design an adaptive fusion block for multiple-pathway 3D MutualNet.
\begin{figure}[!htbp]
    \centering
    \includegraphics[width=0.9\linewidth]{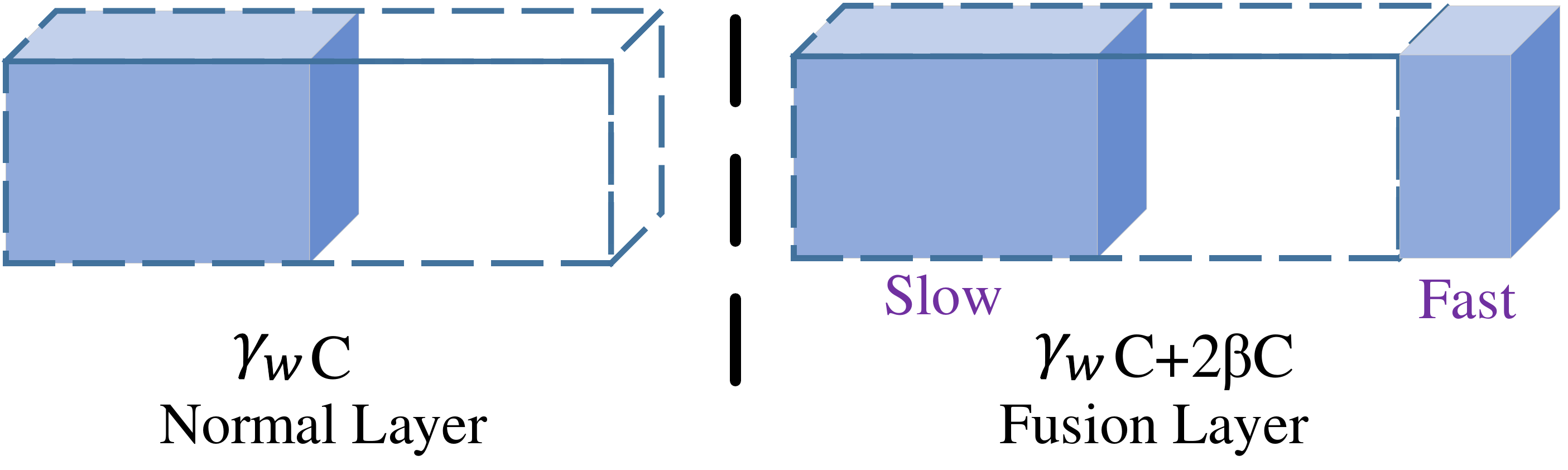}
    \caption{An illustration of adaptive fusion on network channels.}
    \label{fig:fusion}
\end{figure}

Following SlowFast \cite{slowfast}, we denote the feature shape of a standard Slow pathway as $\{T,S^{2},C\}$, where $S$ is the spatial resolution and $C$ is the number of channels. Then the feature shape of adaptive Slow pathway is $\{\gamma_{t}T,(\gamma_{s}S)^{2},\gamma_{w}C\}$. The feature shape of Fast pathway remains $\{\alpha T, S^{2},\beta C\}$ as in \cite{slowfast} ($\alpha, \beta$ are hyperparameters defined in \cite{slowfast}. $\alpha=8,\beta=1/8$ for SlowFast 4$\times$16). Following the settings in \cite{slowfast}, we first perform a 3D convolution of $5\times 1^{2}$ kernel with $2\beta C$ output channels and a stride of $\alpha$. The output feature shape of this convolution layer is $\{T, S^{2}, 2\beta C\}$. To fuse it with the adaptive Slow pathway (whose shape is $\{\gamma_{t}T,(\gamma_{s}S)^{2},\gamma_{w}C\}$), we perform a spatial interpolation and temporal down-sampling to make the output shape $\{\gamma_{t}T,(\gamma_{s}S)^{2},2\beta C\}$. Then the final feature shape after the fusion is $\{\gamma_{t}T,(\gamma_{s}S)^{2},(\gamma_{w}+2\beta)C\}$. As shown in Fig. \ref{fig:fusion}, normal convolution layers of adaptive Slow pathway have $\gamma_{w}C$ channels indexing from the left, while the first convolution layer after each fusion has $\gamma_{w}C+2\beta C$ input channels. The last $2\beta C$ channels are always kept for the output of Fast pathway, while the first $\gamma_{w}C$ channels from the left can vary for each iteration. This operation enforces an exact channel-wise correspondence between the fusion features and the parameters in convolution layers.

\subsection{3D Model inference}
Similar to 2D networks, we need to find the best-performing configuration at each resource constraint. We evaluate different width-spatial-temporal configurations on a validation set. Then  the complexity-configuration table can be obtained by following the steps in Section \ref{sec:2d-inference}. For real deployment, the model and the table need to be maintained, and therefore the memory consumption is essentially the same as a single model.  The model can be adjusted according to the table to meet different resource budgets.

\begin{table}[!t]
    \huge
    \renewcommand{\arraystretch}{1.2}
    \caption{Training cost of MutualNet and independent models of different scales. The backbone is MobileNet v1. $^*$ indicates expected values.}
    \label{tab:2dtrainingcost}
    \centering
    \resizebox{\linewidth}{!}{
    \begin{tabular}{r|c|c|c|c|c|c|c}
    \hline
    MobileNet v1  & \multicolumn{6}{c|}{Independent} & MutualNet \\
    \hline
      Scale & $\times$0.1 & $\times$0.3 & $\times$0.5 & $\times$0.7 & $\times$0.9 & $\times$1.0 & $\times$0.02 $\sim\times$1.0 \\
     \hline
     MFLOPs & 57 & 171 & 284 & 398 & 512 & 569 & 910$^{*}$ \\
     \hline
     Total & \multicolumn{6}{c|}{1991} & 910$^{*}$ \\
     \hline
     Mins/epoch & 1$^{*}$ & 3$^{*}$ & 5$^{*}$ & 7$^{*}$ & 9$^{*}$ & 10 & 20 \\
     \hline
     Total & \multicolumn{6}{c|}{35$^*$} & 20 \\
     \hline
    \end{tabular}
    }
\end{table}
\begin{table}[!t]
    \huge
    \renewcommand{\arraystretch}{1.2}
    \caption{Training cost of MutualNet and independent models of different scales. The backbone is Slow-8$\times$8. $^*$ indicates expected values.}
    \label{tab:3dtrainingcost}
    \centering
    \resizebox{\linewidth}{!}{
    \begin{tabular}{r|c|c|c|c|c|c|c}
    \hline
    Slow-8$\times$8  & \multicolumn{6}{c|}{Independent} & MutualNet \\
    \hline
      Scale & $\times$0.1 & $\times$0.3 & $\times$0.5 & $\times$0.7 & $\times$0.9 & $\times$1.0 & $\times$0.06 $\sim\times$1.0 \\
     \hline
     GFLOPs & 5.5 & 16.4 & 27.3 & 38.2 & 49.1 & 54.5 & 74.8$^{*}$ \\
     \hline
     Total & \multicolumn{6}{c|}{191} & 74.8$^{*}$ \\
     \hline
     Mins/epoch & 5.8$^{*}$ & 17.4$^{*}$ & 29.0$^{*}$ & 40.6$^{*}$ & 52.2$^{*}$ & 58 & 69 \\
     \hline
     Total & \multicolumn{6}{c|}{203$^*$} & 69 \\
     \hline
    \end{tabular}
    }
\end{table}

\section{Complexity of MutualNet Training}\label{sec:complexity}
Since we additionally sample sub-networks during training, MutualNet consumes more computational cost than training a single model. However, we show that the training cost is several times less than training many independent models. In Table \ref{tab:2dtrainingcost}, we measure the theoretical complexity (i.e., FLOPs) and practical wall-clock time of training MutualNet and independent models. The network backbone is MobileNetv1 \cite{mobilenet}. In MutualNet, the width range is [0.25, 1.0]$\times$ and the resolution range is \{224, 192, 160, 128\}. This achieves a dynamic constraint of [13, 569] MFLOPs, which corresponds to a model scale from $\times$0.02 to $\times$1.0. Note that we use 1.0$\times$ to denote the network width coefficient while $\times$1.0 to denote the overall model scale. For independent training, we train a set of models where the smallest one is $\times$0.1 and the stepsize is $\times$0.2. The FLOPs of MutualNet training is estimated by taking the expectation of Eq. \ref{eq:2dreducecost}. The wall-clock time of the full model ($\times$1.0) and MutualNet is measured on an $8\times$2080TI GPU sever with a batch-size of 1024. Other models' training time is estimated by the model FLOPs because the practical time depends on the manner the model is scaled down.  However, it should be higher than the estimated values because the data loading/processing time does not decrease if the model is scaled down.   Table \ref{tab:2dtrainingcost} shows that the training cost of MutualNet is around 2 times of the full-model, but it is much smaller than independently training each model.

\begin{figure*}[!ht]
\centering
    \includegraphics[width=\textwidth]{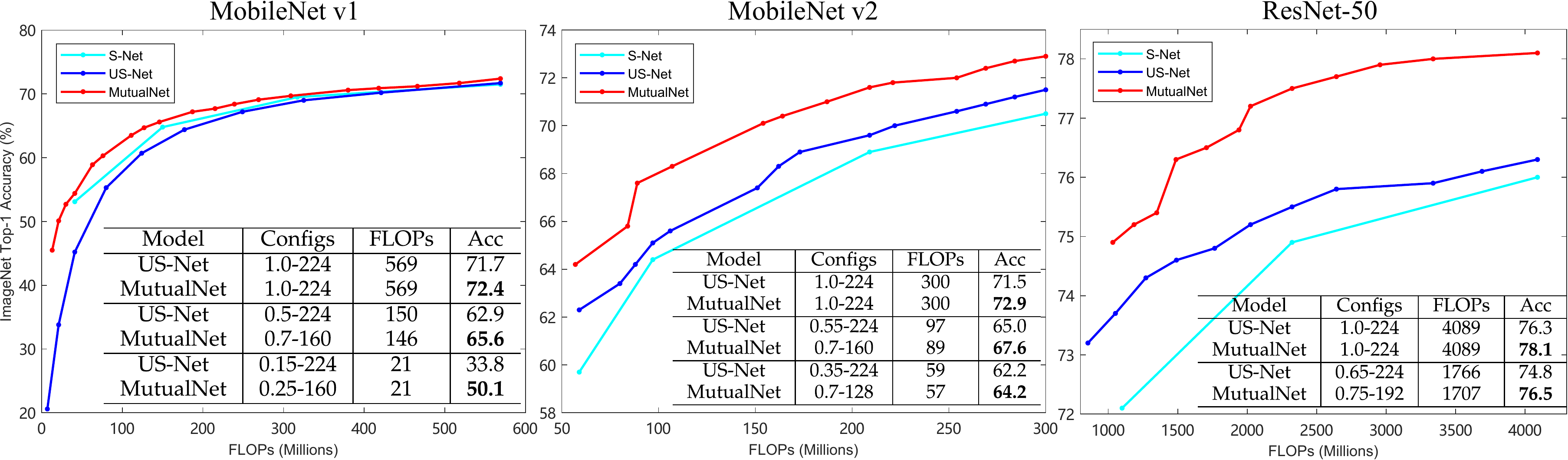}
    \caption{Comparisons of Accuracy-FLOPs curves of MutualNet, S-Net and US-Net. In the tables, we compare some points on the curves by their configurations, the corresponding FLOPs and accuracy.}
    \label{fig:2d_slim_IN1k}
\end{figure*}

\begin{figure*}[!ht]
\centering
    \includegraphics[width=\textwidth]{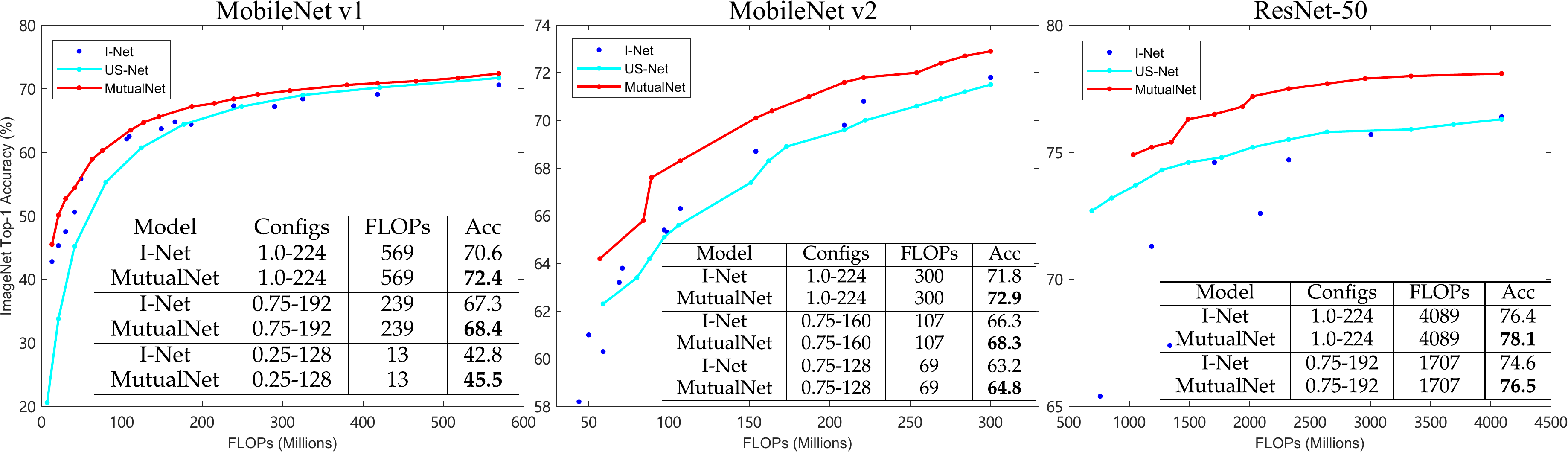}
    \caption{Comparisons of Accuracy-FLOPs curves of MutualNet, US-Net and I-Net.}
    \label{fig:2d_individual_IN1k}
\end{figure*}

We also evaluate the training cost on 3D networks (Slow-8$\times$8 \cite{slowfast}) in Table \ref{tab:3dtrainingcost}. The training time is measured on an $8\times$2080TI GPU sever with a batch-size of 64. In MutualNet, the width range is [0.63, 1.0]$\times$, spatial resolution is \{142, 178, 224\} and temporal resolution is \{3, 5, 8\}. This achieves a model scale from $\times$0.06 to $\times$1.0. We also evaluate a group of independent models from $\times$0.1 to $\times$1.0. We can see that although the theoretical complexity of MutualNet is about 1.4 times of the full-model, the wall-clock time is only slightly higher. This is because the data loading/processing is time-consuming in 3D networks, the additional cost introduced by sub-networks is therefore not that significant. {\em In summary, MutualNet saves time compared to independently training several models, and it only needs to deploy one model to meet dynamic resource constraints during inference.}

\section{Experiments}\label{sec:experiments}
We conduct extensive experiments to evaluate the effectiveness of our proposed method. On 2D network structures, we first present our results on ImageNet \cite{imagenet} classification to illustrate the effectiveness of MutualNet. Next, we conduct extensive ablation studies to analyze the mutual learning scheme. Finally, we apply MutualNet to transfer learning datasets and COCO \cite{coco} object detection and instance segmentation to demonstrate its robustness and generalization ability. On 3D network structures, we conduct experiments on three video datasets following the standard evaluation protocols. We first evaluate our method on Kinetics-400 \cite{kay2017kinetics} for action classification. Then we transfer the learned representations to Charades \cite{charade} action classification and AVA \cite{gu2018ava} action detection. 

\subsection{Evaluation on ImageNet Classification}
\label{sec:evalImageNet}
We compare MutualNet with SlimmableNet (S-Net \cite{slimnet} and US-Net \cite{usnet}) and independently-trained networks on the ImageNet dataset. 
We evaluate our framework on three popular network structures, MobileNetv1 \cite{mobilenet}, MobileNetv2 \cite{sandler2018mobilenetv2} and ResNet-50 \cite{resnet}. 

{\bf Implementation Details.} We follow the settings in SlimmableNet and make the comparison under the same dynamic FLOPs constraints: [13, 569] MFLOPs on MobileNetv1, [57, 300] MFLOPs on MobileNetv2 and [660, 4100] MFLOPs on ResNet-50. The input image resolution is randomly picked from \{224, 192, 160, 128\} unless specified. We use width scale [0.25, 1.0]$\times$ on MobileNetv1, [0.7, 1.0]$\times$ on MobileNetv2 and [0.7, 1.0]$\times$ on ResNet-50. The width lower bound is slightly higher than that in SlimmableNet because we perform multi-dimension trade-off during training. The other training settings are the same as SlimmableNet.

\begin{figure*}[t]
    \centering
    \includegraphics[width=0.9\textwidth]{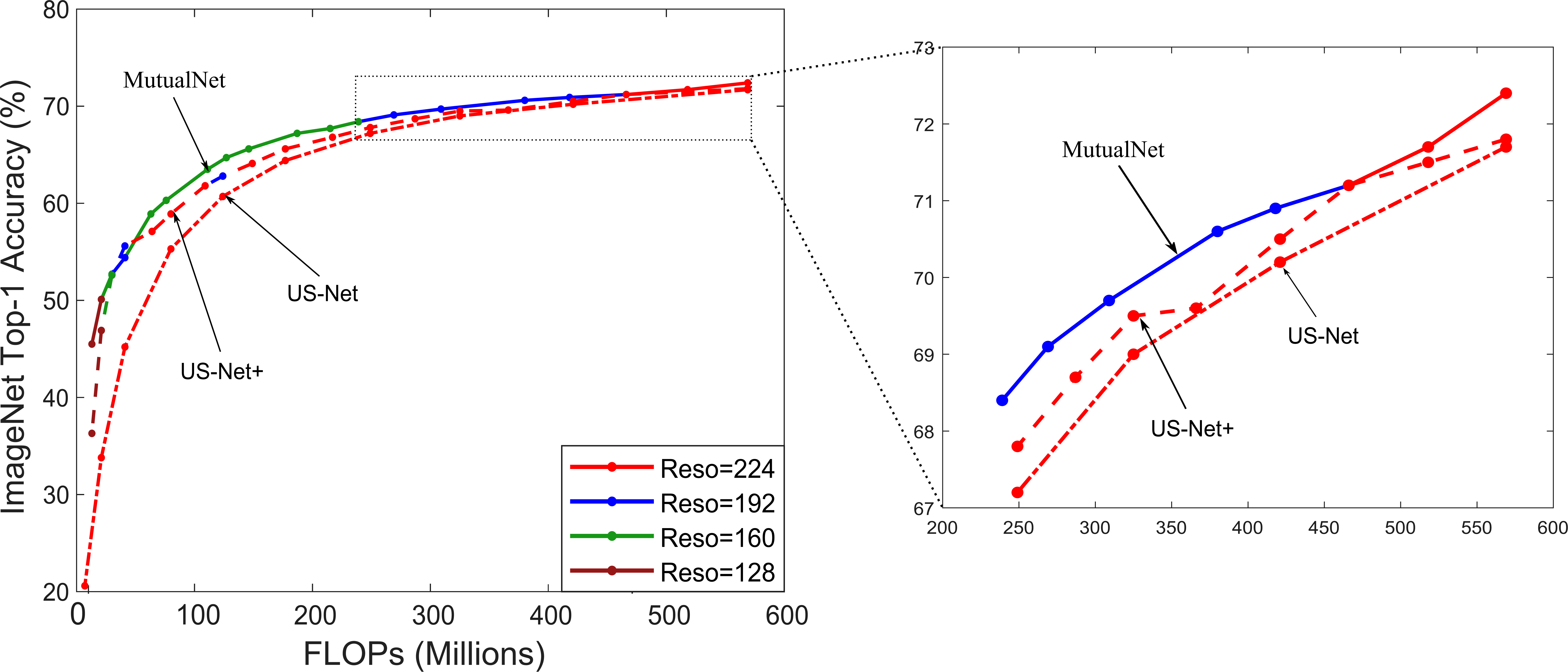} 
    \caption{The width-resolution trade-offs at different resource constraints. The Accuracy-FLOPs curves are based on MobileNet v1 backbone. We highlight the selected resolution under different FLOPs with different colors. For example, the solid green line indicates that when the constraint range is [41, 215] MFLOPs, our method constantly selects input resolution 160 but reduces the width to meet the resource constraint. Best viewed in color.}
    \label{fig:Ours-US-Net+}
\end{figure*}
    
{\bf Comparison with SlimmableNet.} The Accuracy-FLOPs curves are shown in Fig. \ref{fig:2d_slim_IN1k}. We can see that our method consistently outperforms S-Net and US-Net on MobileNetv1, MobileNetv2 and ResNet-50 backbones. Specifically, we achieve significant improvements under small computation costs. This is because our framework considers both network width and input resolution and can find a better balance between them. For example, on MobileNet v1 backbone, if the resource constraint is 150 MFLOPs, US-Net has to reduce the width to $0.5\times$ given its constant input resolution $224$, while MutualNet can meet this budget by a balanced configuration of ($0.7\times$ - 160), leading to a better accuracy (65.6\% (Ours) vs. 62.9\% (US-Net) as listed in the table of Fig. \ref{fig:2d_slim_IN1k}). Also, our framework is able to learn multi-scale representations as demonstrated in Section \ref{sec:2d-analysis}, which further boost the performance of each sub-network. We can see that even for the same configuration (e.g., 1.0$\times$-224) our approach clearly outperforms US-Net, i.e., 72.4\% (Ours) vs. 71.7\% (US-Net) on MobileNet v1, 72.9\% (Ours) vs. 71.5\% (US-Net) on MobileNet v2, and 78.1\% (Ours) vs. 76.3\% (US-Net) on ResNet-50. (Fig. \ref{fig:2d_slim_IN1k}).

{\bf Comparison with Independently Trained Networks.} We compare the performance of MutualNet and US-Net with independently-trained networks (\textbf{denoted by I-Net}) under different width-resolution configurations in Fig. \ref{fig:2d_individual_IN1k}. In I-Net, the resolutions are selected from \{224, 192, 160, 128\}. Width are selected from \{$1.0\times$, $0.75\times$, $0.5\times$, $0.25\times$\} on MobileNet v1\&v2 and \{$1.0\times$\, $0.75\times$\} on ResNet-50. From Fig. \ref{fig:2d_individual_IN1k} we can see that US-Net only achieves comparable (in many cases worse) performance compared to I-Net, while MutualNet consistently outperforms US-Net and I-Net on three backbones. Even at the same width-resolution configuration, which  may not be the best configuration,  
MutualNet can achieve much better performance than I-Net. This demonstrates that MutualNet not only finds the better width-resolution balance but also learns stronger representations by the mutual learning scheme.


{\bf Balanced Width-Resolution Configuration via Mutual Learning.} 
One may apply different resolutions to US-Net \textit{during inference} to yield improvement over the original US-Net. However, this way the optimal width-resolution balance can be achieved due to lack of width-resolution mutual learning. In one  experiment, we evaluate US-Net at width scale [0.25, 1.0]$\times$ with input resolutions \{224, 192, 160, 128\} and denote this improved model as \textbf{US-Net+}. In Fig. \ref{fig:Ours-US-Net+}, we plot the Accuracy-FLOPs curves of our method and US-Net+ based on MobileNet v1 backbone, and highlight the selected input resolutions with different colors. As we decrease the FLOPs ($569\rightarrow468$ MFLOPs), MutualNet first reduces network width to meet the constraint while keeping the 224$\times$224 resolution (red line in Fig. \ref{fig:Ours-US-Net+}). After 468 MFLOPs, MutualNet selects lower input resolution (192) and then continues reducing the width to meet the constraint. On the other hand, US-Net+ cannot find such balance. It always slims the network width and uses the same (224) resolution as the FLOPs decreasing until it goes to really low. This is because US-Net+ does not incorporate input resolution into the learning framework. \textit{Simply applying different resolutions during inference cannot achieve the optimal width-resolution balance.}

{\bf Comparison with EfficientNet.}
EfficientNet \cite{tan2019efficientnet} acknowledges the importance of balancing among network width, depth and resolution. But they are considered as independent factors. The authors use grid search over these three dimensions and train each model configuration independently to find the optimal one under certain constraint, while MutualNet incorporates width and resolution in a unified training framework. To show the benefits of the mutual learning shceme, we compare MutualNet with the best model scaling that EfficientNet finds for MobileNet v1 at 2.3 BFLOPs (scale up baseline by $\times$4.0). To cover this model scale we scale up MutualNet by using a width range of [$1.0\times,2.0\times$], and select resolutions from \{224, 256, 288, 320\}. This makes MutualNet executable in the range of [0.57, 4.5] BFLOPs. We pick the best performing width-resolution configuration at 2.3 BFLOPs. The results are compared in Table \ref{tab:diffeffientnet}. Although EfficientNet claims to find the optimal scaling compound, its performance is much worse than MutualNet. This is because EfficientNet fails to leverage the information in other configurations, while MutualNet captures multi-scale representations for each model configuration thanks to the width-resolution mutual learning.

\begin{table}[t]
\renewcommand{\arraystretch}{1.2}
\caption{Comparison with EfficienNet to scale up MobileNetv1 by $\times$4 on ImageNet. $d$: depth, $w$: width, $r$: resolution.}
\begin{center}
\resizebox{\linewidth}{!}{
\begin{tabular}{c|c|c|c}
    \hline
     Model& Best Model & FLOPs& Top-1 Acc  \\
     \hline
     EfficientNet \cite{tan2019efficientnet}& $d=1.4, w=1.2, r=1.3$& 2.3B& 75.6\% \\
     MutualNet& $w=1.6, r=1.3$& 2.3B& {\bf 77.1\%} \\
     \hline
\end{tabular}
}
\end{center}
\label{tab:diffeffientnet}
\end{table}

{\bf {Comparison with Multi-scale Data Augmentation.}}
In multi-scale data augmentation, 
the network may take images of different resolutions in different training iterations. 
But within each iteration, the network weights are still optimized in the direction of  the same resolution. In contrast, our method randomly samples several sub-networks which share weights with each other. Since sub-networks can select different image resolutions, the weights are optimized in the direction of mixed resolution  in each iteration as illustrated in Fig. \ref{mutual}. This enables each sub-network to effectively learn multi-scale representations from both network width and resolution. To validate the superiority of our mutual learning scheme, we apply multi-scale data augmentation to I-Net and US-Net and explain the difference with MutualNet.

\textit{I-Net + Multi-scale data augmentation.} We train MobileNetv2 ($1.0\times$ width) with multi-scale images. To have a fair comparison, input images are randomly sampled from scales \{224, 192, 160, 128\} and the other settings are the same as MutualNet. As shown in Table \ref{tab:mnetmultiscale}, multi-scale data augmentation only marginally improves the baseline (MobileNetv2) while MutualNet (MobileNetv2 backbone) clearly outperforms both of them by considerable margins.

\begin{table}[t]
\renewcommand{\arraystretch}{1.2}
\caption{Comparison between MutualNet and multi-scale data augmentation.}
\begin{center}
\begin{tabular}{c|c}
    \hline
     Model& ImageNet Top-1 Acc  \\
     \hline
     MobileNet v2 (1.0$\times$ - 224) - Baseline & 71.8\% \\
     Baseline + Multi-scale data augmentation & 72.0\% \\
     MutualNet (MobileNet v2 backbone) & {\bf72.9\%}\\
     \hline
\end{tabular}
\end{center}
\label{tab:mnetmultiscale}
\end{table}

\begin{figure}[t]
\centering
    \includegraphics[width=0.95\linewidth]{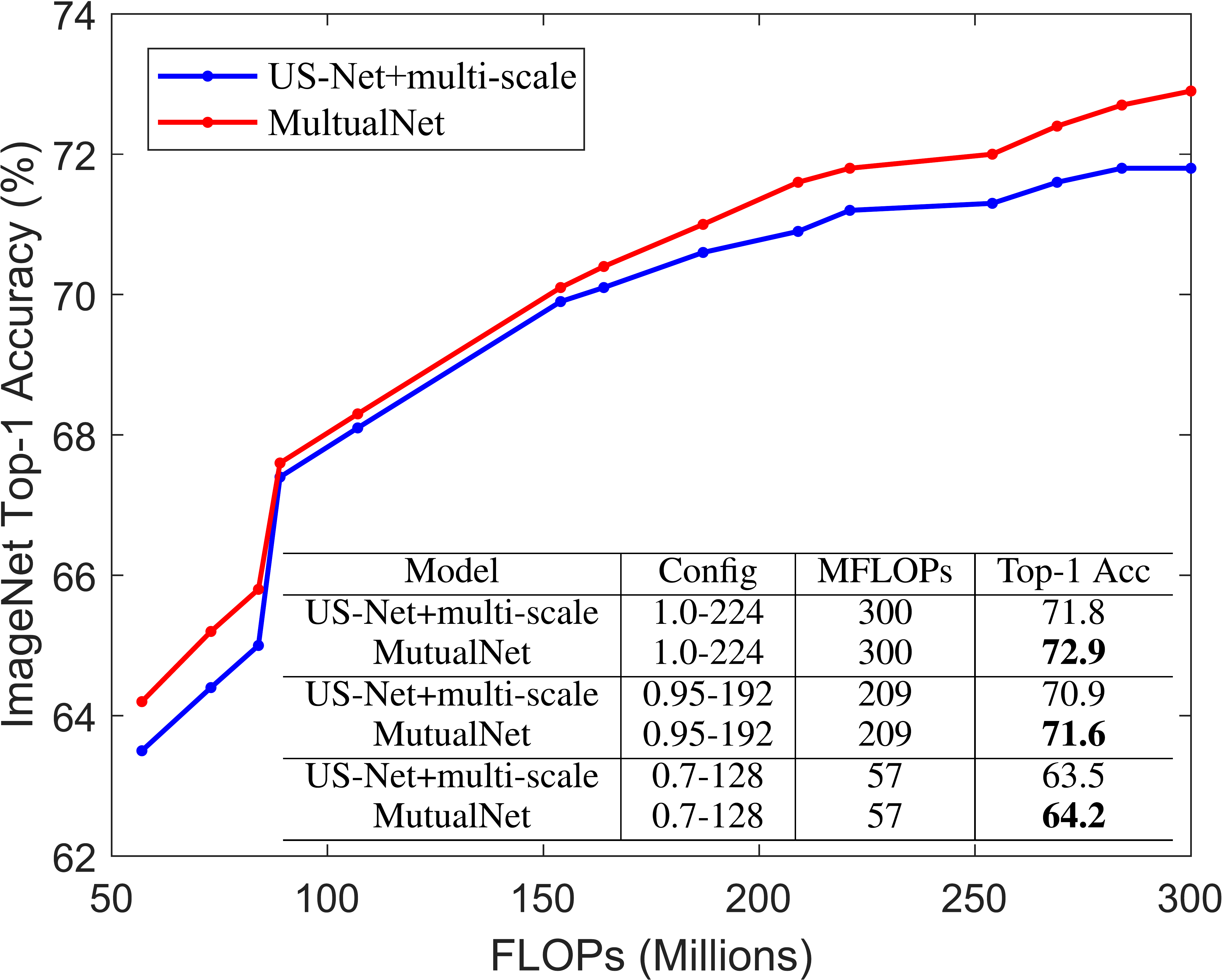} 
    \caption{MutualNet and US-Net + multi-scale data augmentation.}
    \label{fig:USmultiscale}
\end{figure}

\begin{figure}[t]
    \centering
    \includegraphics[width=0.95\linewidth]{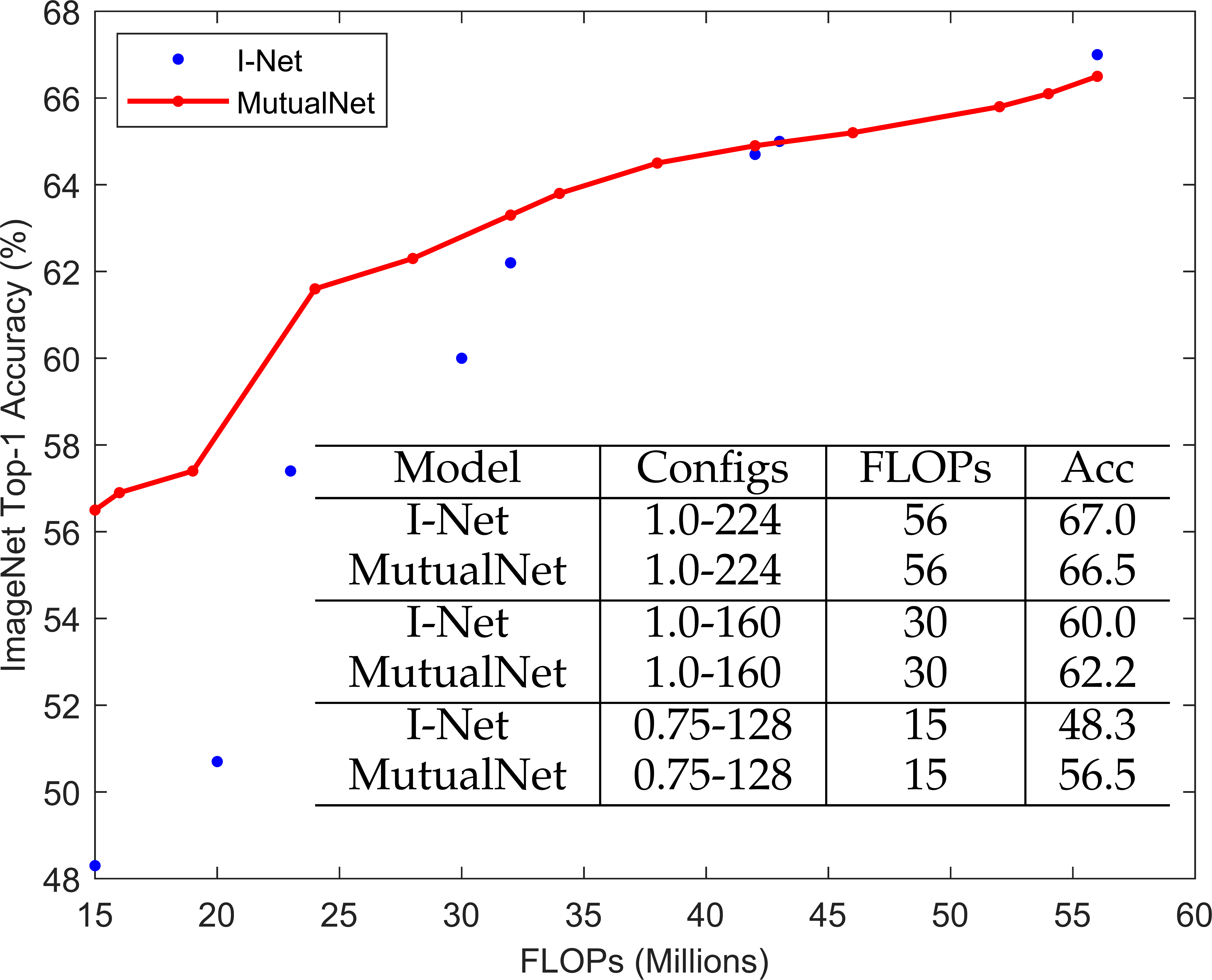}
    \caption{Comparison of independently-trained networks (I-Net) and MutualNet with the MobileNetV3 backbone.}
    \label{fig:ind_mnetv3}
    \end{figure}

\begin{figure}[t]
\centering
\includegraphics[width=0.95\linewidth]{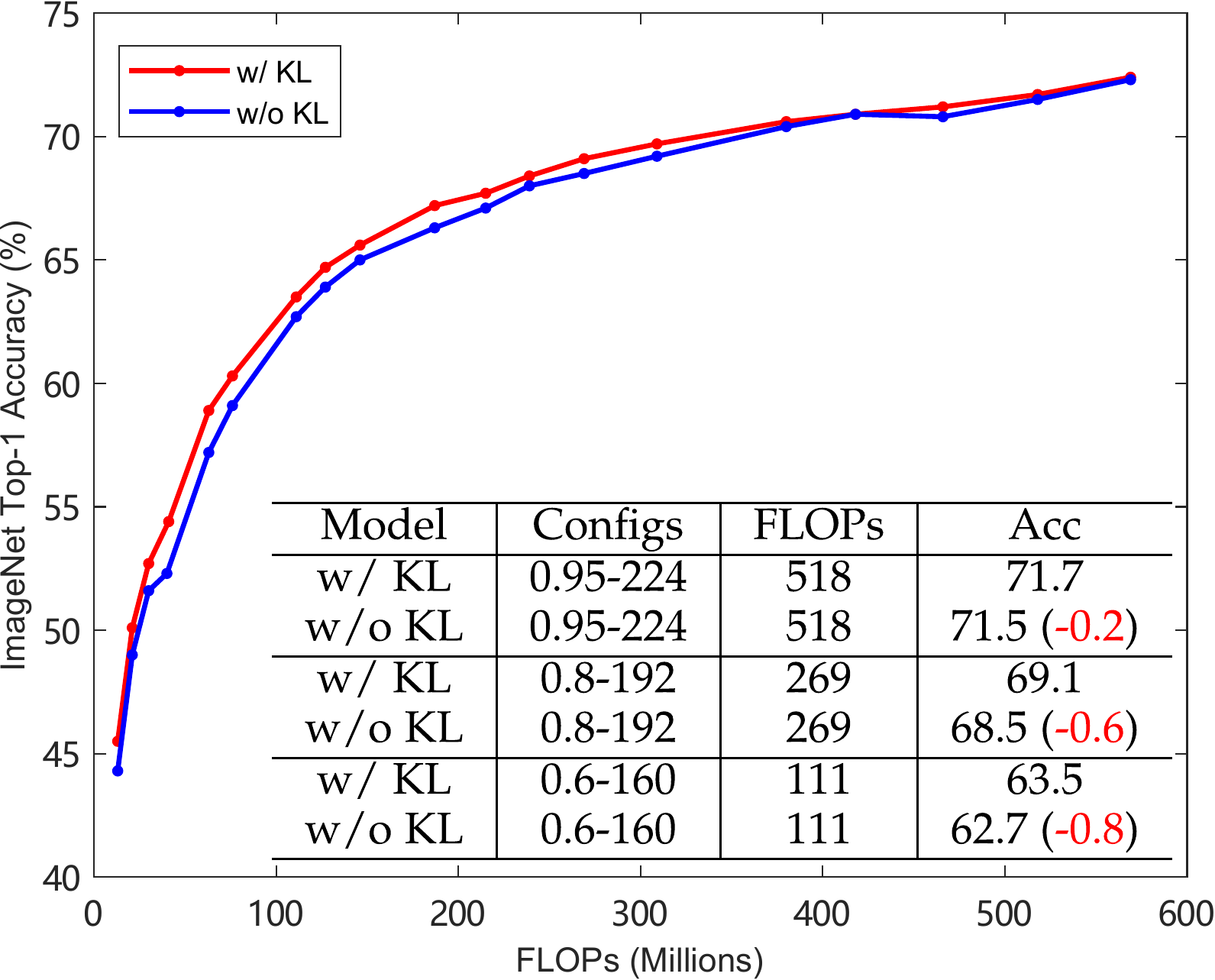} 
    \caption{Contribution of the KLDiv loss to the overall performance.}
    \label{fig:kleffect}
\end{figure}

\begin{figure}[t]
\centering
\includegraphics[width=0.95\linewidth]{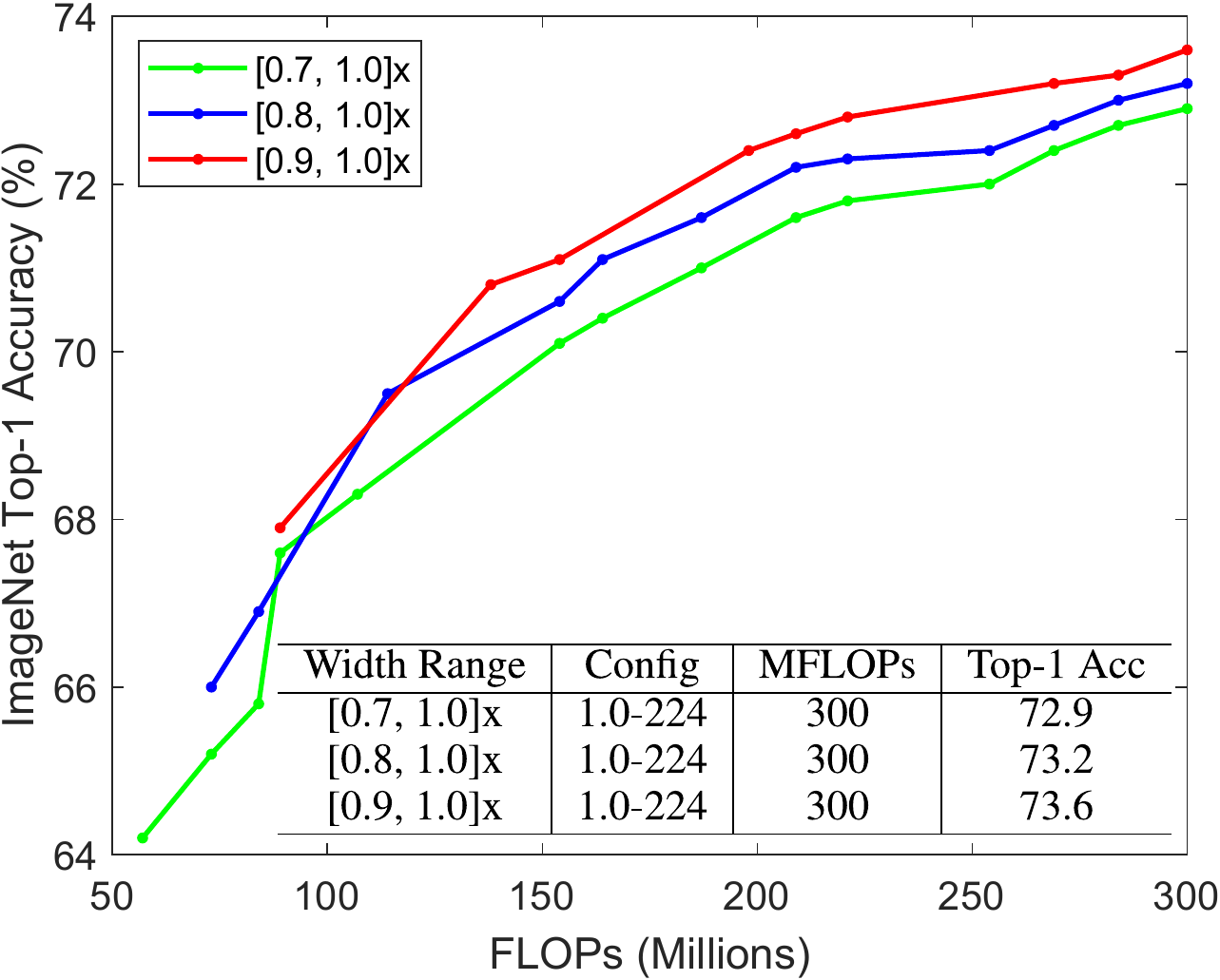} 
    \caption{Accuracy-FLOPs curves of different width lower bounds.}
    \label{fig:lowerbound}
\end{figure}

\begin{figure}[t]
\centering
    \includegraphics[width=0.95\linewidth]{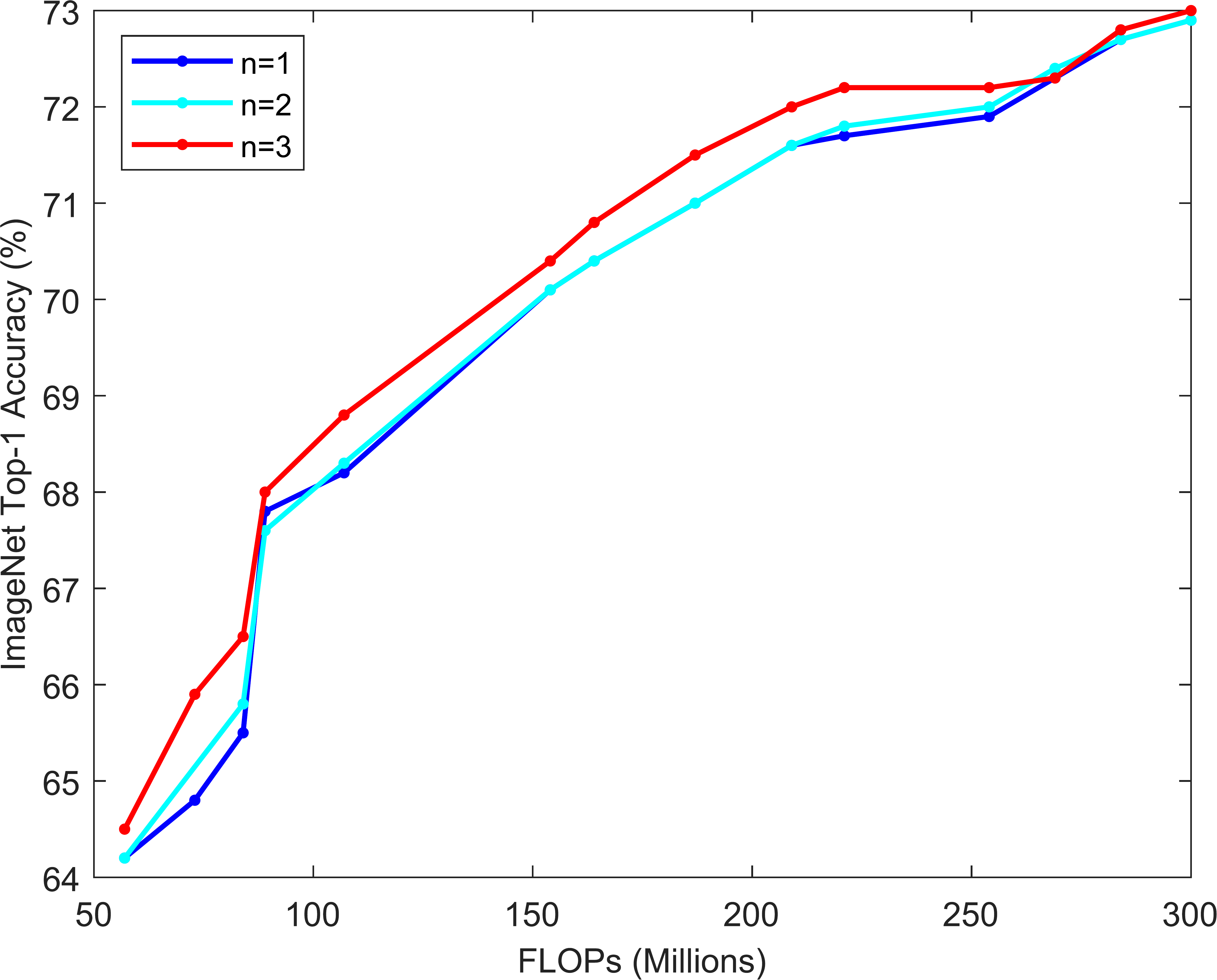}
    \caption{The effect of the number of randomly sampled sub-networks during training. The backbone network is MobileNetV2. $n=1$ means one random width sub-network is sampled with the full-network and the smallest sub-network ($\gamma_w=0.7$).}
    \label{fig:number_subnet}
\end{figure}

\begin{figure*}[!ht]
    \centering
    \includegraphics[width=0.97\linewidth]{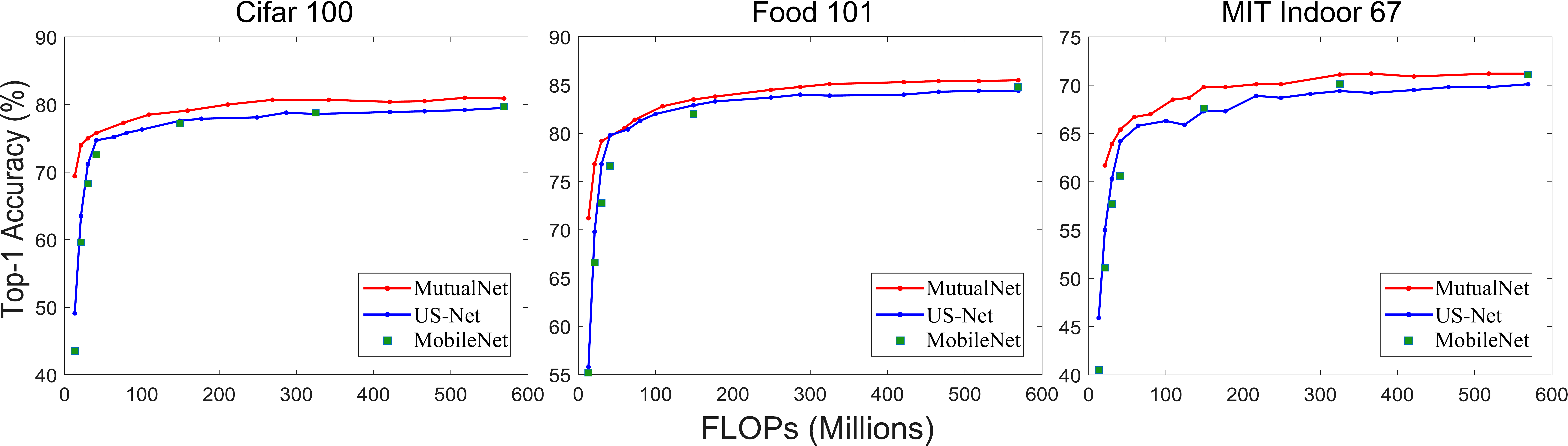} 
    \caption{Accuracy-FLOPs curves of different methods on different transfer learning datasets. MobileNet is trained independently at different model configurations.}
    \label{fig:transfer}
\end{figure*}

\textit{US-Net + Multi-scale data augmentation.} Different from our framework which feeds different scaled images to different sub-networks, in this experiment, we \textit{randomly} choose a scale from \{224, 192, 160, 128\} and feed the \textit{same} scaled image to all sub-networks in each iteration. That is, each sub-network takes the same image resolution. In this way, the weights are still optimized towards a single resolution direction in each iteration. {For example, as illustrated in Fig. \ref{mutual}, the gradient of the sub-network $0.4 \times$ in MutualNet is $\frac{\partial l_{W_{0:0.4},I_{R=128}} + \partial l_{W_{0:0.8},I_{R=192}}}{\partial W_{0:0.4}}$, while in US-Net + multi-scale it would be $\frac{\partial l_{W_{0:0.4},I_{R=128}} + \partial l_{W_{0:0.8},I_{R=128}}}{\partial W_{0:0.4}}$. With more sub-networks and input scales involved, the difference between their gradient flows becomes more distinct. As shown in Fig. \ref{fig:USmultiscale}, our method clearly outperforms  \textit{US-Net + multi-scale data augmentation} over the entire FLOPs spectrum. This experiment is based on MobileNetv2 with the same settings as in Sec. \ref{sec:evalImageNet}.} \textit{These experiments demonstrate that the improvement comes from our mutual learning scheme rather than the multi-scale data augmentation.}

\textbf{Combine with Dynamic Blocks.} As reviewed in Section \ref{sec:relatedwork}, there is a category of dynamic networks where the adaptive weights are determined by the input. We show that MutualNet can be directly combined with these networks by applying our method to MobileNetV3 \cite{mobilenetv3}, where the dynamic blocks are implemented by Squeeze and Excitation (SE) \cite{senet}. When sampling sub-networks, we multiply the width factor to both the backbone network and SE block. Then the full-network and sub-networks can be trained in the same way as other models. We compare MutualNet and independently trained MobileNetV3 (denoted by \textbf{I-Net}) at different configurations in Fig. \ref{fig:ind_mnetv3}. Note that the results of MobileNetV3 are reproduced by us since we could not strictly follow the settings in the original paper (the authors trained MobileNetV3 on 4$\times$4 TPU Pod with a batch size of 4096). We train the network on an 8-GPU server for 150 epochs with a batch size of 1024. The initial learning rate is 0.4 with cosine decay schedule. Our reproduced performance is 0.4\% lower than that in the original paper. MutualNet is trained on the same codebase with the same settings. As shown in Fig. \ref{fig:ind_mnetv3}, MutualNet does not outperforms independently-trained MobileNetV3 at large model configurations. We conjecture this is caused by the SE block in MobileNetV3. In MutualNet, sub-networks and the full-network share the same SE block, but their channel attention could be different (sub-networks do not have some channels). Fitting the channel attention to sub-networks may hurt the attention of the full-network, thus leading to decreased performance. But at smaller configurations, MutualNet is significantly better. Although the overall improvement may not be as significant as on other backbones, MutualNet still has the advantage of covering a wide range of resource constraints by a single model, which makes it easier to deploy on resource-constrained devices. Still, we think that investigating the effective combination of MutualNet with other dynamic inference methods is an interesting problem. We leave further study on this for future work.

\subsection{Abalation study}\label{sec:ablation}
\textbf{Effects of KL Divergence.} During training, we leverage the full-network to supervise sub-networks to enhance knowledge transfer. Here we study how much does the KL Divergence loss contribute to the overall performance. As shown in Fig. \ref{fig:kleffect}, \textit{w/ KL} is the original training process and \textit{w/o KL} denotes that the sub-networks are supervised by the ground truth labels. The difference is very marginal where the largest gap is less than 1\%. This demonstrates that the KL Divergence loss does benefit the performance, but the main contribution is coming from the mutual learning scheme as explained in Section \ref{sec:2d-analysis}.

{\bf Effects of Width Lower Bound.} The dynamic constraint is affected by the width lower bound. To study its effects, we conduct experiments with three different width lower bounds (0.7$\times$, 0.8$\times$, 0.9$\times$) on MobileNetv2. The results in Fig. \ref{fig:lowerbound} show that a higher lower bound gives better overall performance, but the dynamic range is narrower. 
One interesting observation is that the performance of the full-network (1.0$\times$-224) is also largely improved as the width lower bound increases from 0.7$\times$ to 0.9$\times$. This property is not observed in US-Net. We attribute this to the robust and well-generalized multi-scale representations which can be effectively re-used by the full-network, while in US-Net, the full-network cannot effectively benefit from 
sub-networks.

\begin{table}[t]
\renewcommand{\arraystretch}{1.2}
\centering
  \caption{Comparisons of the Top-1 Accuracy (\%) of MutualNet and state-of-the-art techniques for boosting a single network.}
  \begin{tabular}{c|c|c|c}
    \hline
     Method & Cifar-10 & Cifar-100 & ImageNet  \\
     \hline
     Baseline \cite{wideresnet, resnet} & 96.1 & 81.2 & 76.5 \\
     \hline 
     Cutout \cite{cutout}& 96.9 & 81.6 & 77.1 \\
     SENet \cite{senet}& / & / & 77.6 \\
     AutoAug \cite{cubuk2019autoaugment}& \textbf{97.4} & 82.9 & 77.6 \\
     ShakeDrop \cite{yamada2019shakedrop} & 95.6 & 81.7 & 77.5 \\
     Mixup \cite{zhang2017mixup} & 97.3 & 82.5 & 77.9  \\
     MutualNet & 97.2 & \textbf{83.8} & \textbf{78.6} \\
     \hline
  \end{tabular}
  \label{tab:boostsingle-single}
\end{table}

\begin{figure*}[t]
    \centering
    \includegraphics[width=0.9\textwidth]{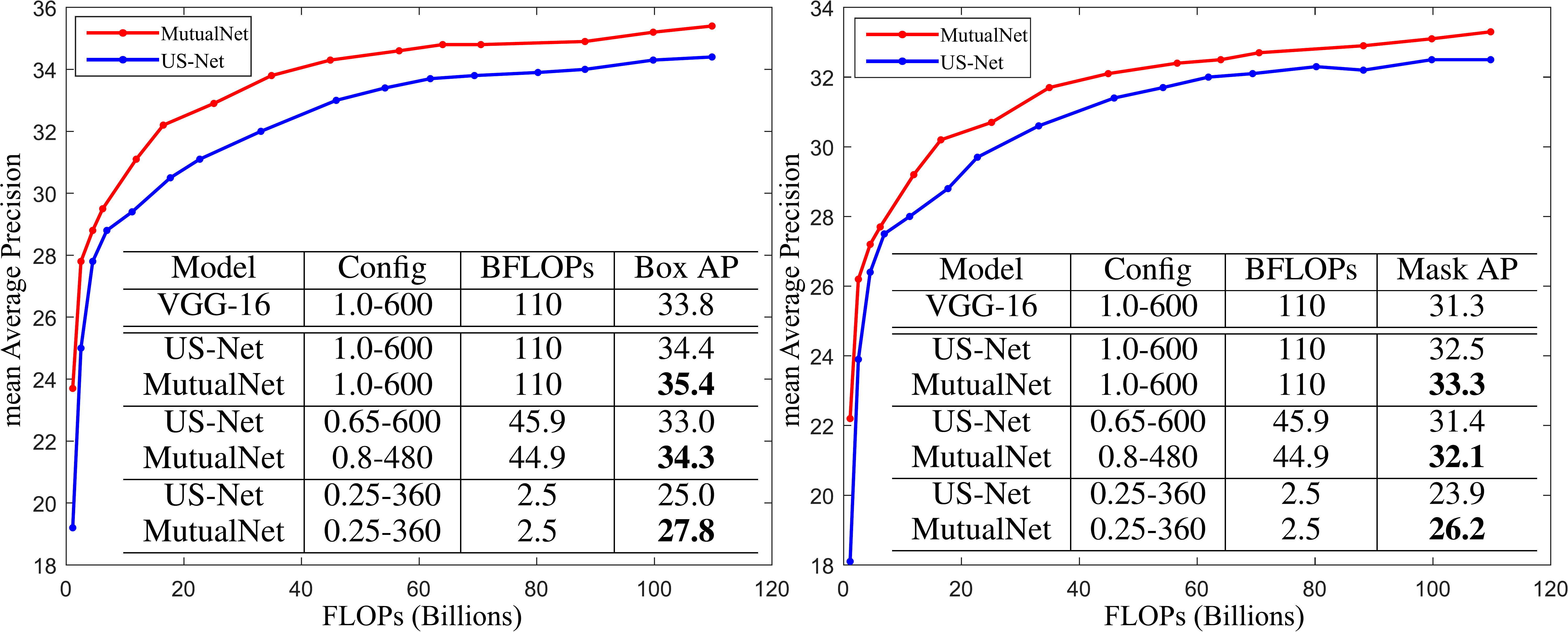}
    \caption{mAP-FLOPs curves of MutualNet and US-Net on object detection (left) and instance segmentation (right). The results are based on Mask-RCNN. All models follow the same training settings.}
    \label{fig:det}
\end{figure*}

\begin{figure*}[t]
    \centering
    \includegraphics[width=0.9\textwidth]{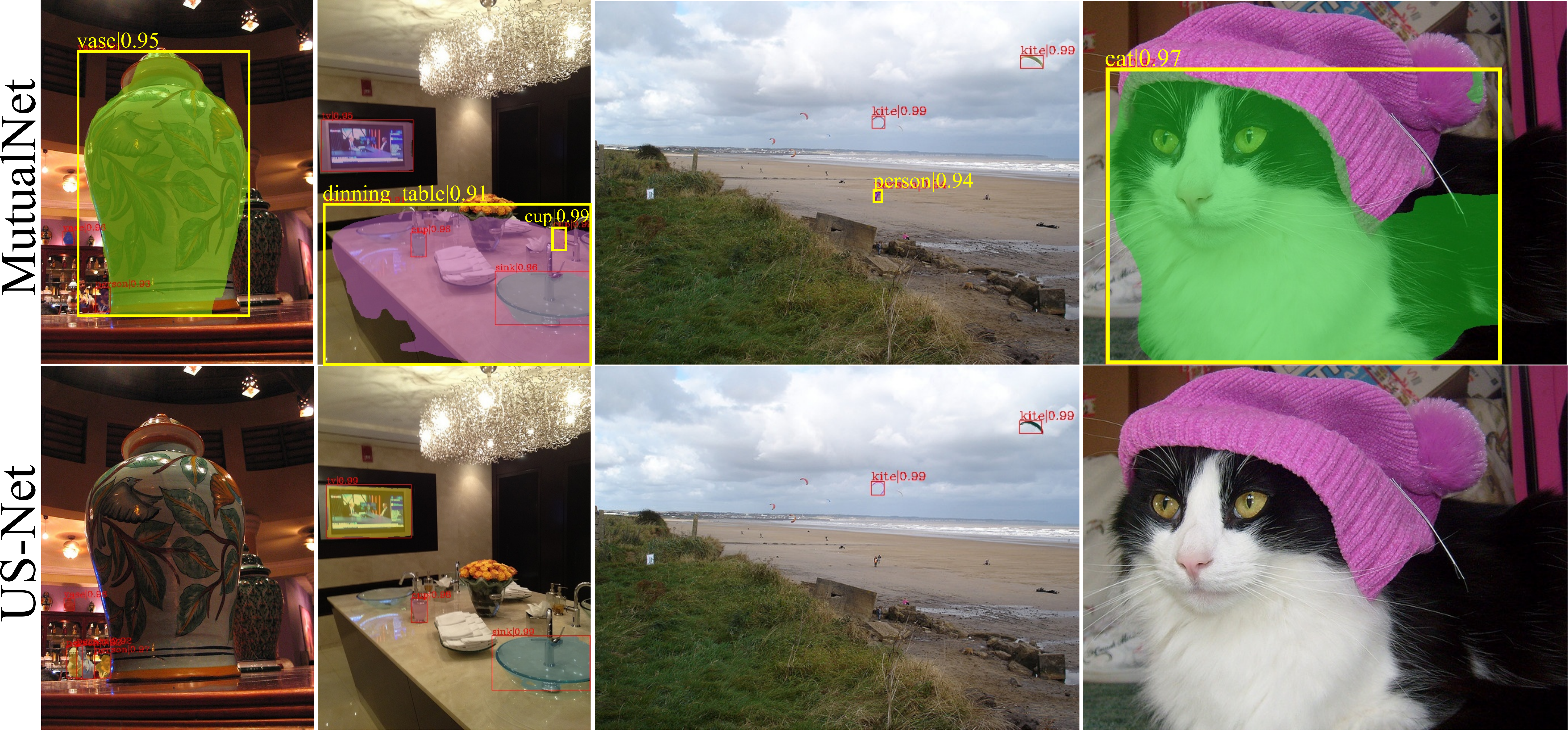}
    \caption{Visualization examples of MutualNet and US-Net on object detection and instance segmentation. Detection and segmentation results are demonstrated by bounding boxes and masks respectively. To facilitate comparison, we use \textbf{yellow boxes} to highlight the objects that MutualNet detects but US-Net fails. [Best viewed with zoom-in.]}
    \label{fig:detvis}
\end{figure*}


{\bf Effect of the number of sub-networks.} Fig. \ref{fig:number_subnet} shows the performance with different numbers of random sub-networks. The backbone network is MobileNetV2. We can see that a larger number of sub-networks could slightly improve the performance over the whole FLOPs spectrum, but it will also increase the training cost. The results also show that even one sub-network can achieve respectable results.

{\bf Boosting Single Network Performance.} As discussed above, the performance of the full-network is greatly improved as we increase the width lower bound. Therefore, we can apply MutualNet to improve the performance of a single full network \textbf{if dynamic budgets is not the concern}. We compare our method with the popular performance-boosting techniques (e.g., AutoAugmentation (AutoAug) \cite{cubuk2019autoaugment}, SENet \cite{senet} and Mixup \cite{zhang2017mixup} etc.) to show its superiority. We conduct experiments using WideResNet-28-10 \cite{wideresnet} on Cifar-10 and Cifar-100 \cite{cifar} and ResNet-50 \cite{resnet} on ImageNet \cite{imagenet}. MutualNet adopts the width range [0.9, 1.0]$\times$ as it achieves the best-performed full-network in Fig. \ref{fig:lowerbound}. The resolution is sampled from \{32, 28, 24, 20\} on Cifar-10 and Cifar-100 and \{224, 192, 160, 128\} on ImageNet. WideResNet is trained for 200 epochs following \cite{wideresnet}. ResNet is trained for 120 epochs. The results are compared in Table \ref{tab:boostsingle-single}. Surprisingly, MutualNet achieves substantial improvements over other techniques even though it is designed to achieve dynamic models. Note that MutualNet is model-agnostic and is as easy as regular training process, so it can take advantage of state-of-the-art network structures and data augmentation techniques.

\subsection{Transfer Learning}
To evaluate the representations learned by our method, we further conduct experiments on three popular transfer learning datasets, Cifar-100 \cite{cifar}, Food-101 \cite{food101} and MIT-Indoor67 \cite{MITindoor67}. Cifar-100  is for superordinate-level object classification, Food-101 is  for fine-grained classification and MIT-Indoor67 is for scene classification. Such  a large variety of datasets can strongly demonstrate the robustness of the learned representations. We compare our approach with US-Net and MobileNetv1. We fine-tune ImageNet pre-trained models
with a batch size of 256, initial learning rate of 0.1 with cosine decay schedule and a total of 100 epochs. Both MutualNet and US-Net are trained with width range [0.25, 1.0]$\times$ and tested with resolutions from \{224, 192, 160, 128\}.
The results are shown in Fig. \ref{fig:transfer}. Again, our MutualNet achieves consistently better performance compared to US-Net and MobileNet. This verifies that MutualNet is able to learn well-generalized representations.

\subsection{Object Detection and Instance Segmentation}
We also evaluate our method on COCO object detection and instance segmentation \cite{coco}. The experiments are based on Mask-RCNN-FPN \cite{maskrcnn, fpn} and MMDetection \cite{mmdetection} toolbox on VGG-16 \cite{vgg} backbone. We first pre-train VGG-16 on ImageNet following US-Net and MutualNet respectively. Both methods are trained with width range [0.25, 1.0]$\times$. Then we fine-tune the pre-trained models on COCO. The feature pyramid network (FPN) neck and detection head are shared among different sub-networks. For simplicity, each sub-network is trained with the ground truth. The other training procedures are the same as training ImageNet classification. Following common settings in object detection, US-Net is trained with image resolution $1000\times600$. Our method randomly selects resolutions from $1000 \times \{600, 480, 360, 240\}$. All models are trained with 2$\times$ schedule for better convergence and tested with different image resolutions. The mean Average Precision (AP at IoU=0.50:0.05:0.95) are presented in Fig. \ref{fig:det}. These results reveal that our MutualNet significantly outperforms US-Net under all resource constraints. Specifically, for the full network (1.0$\times$-600), MutualNet significantly outperforms both US-Net and independent network. This again validates the effectiveness of our width-resolution mutual learning scheme. Fig. \ref{fig:detvis} provides some visual examples which reveal that MutualNet is more robust to small-scale and large-scale objects than US-Net.

\subsection{Evaluation on 3D networks}
To the best of our knowledge, we are the first to achieve adaptive 3D networks. So we only compare our method with independently-trained networks. We conduct experiments based on Slow/SlowFast \cite{slowfast} and X3D \cite{x3d} backbones, which are state-of-the-art 3D network structures. Following previous works \cite{slowfast, x3d}, we evaluate the method on the following three video datasets.

\begin{figure*}[!htbp]
    \centering
    \includegraphics[width=\linewidth]{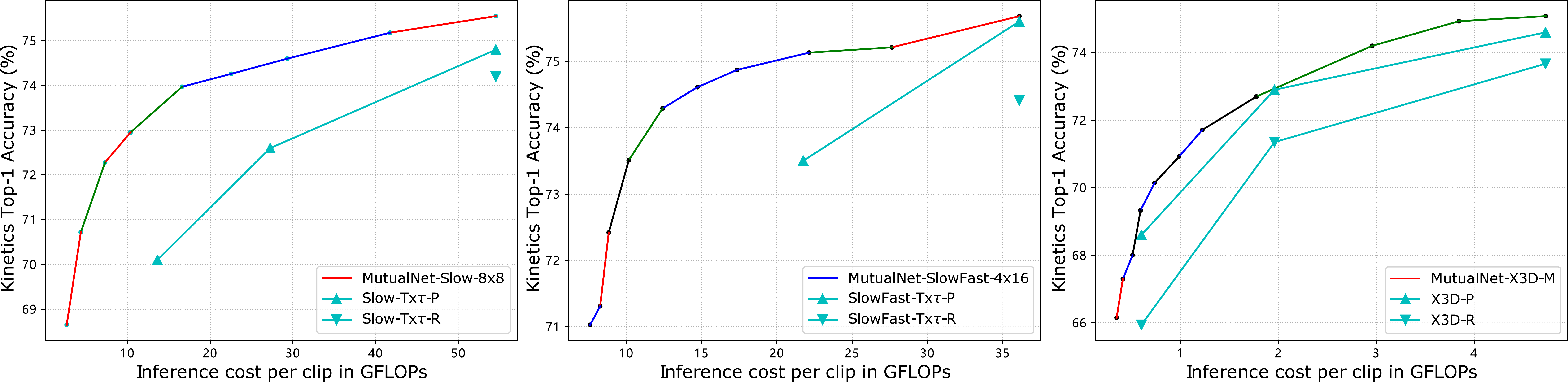}
    \caption{Comparison of MutualNet and its independently trained counterparts under different computational constraints for video action recognition. Both the results in the corresponding papers and our reproduced results are reported for a fair comparison. Our reproduced results are lower because of lack of training data.}
    \label{fig:3d-kinetics}
\end{figure*}

\textbf{Kinetics-400} \cite{kay2017kinetics} is a large scale action classification dataset with $\sim$240k training videos and 20k validation videos trimmed as 10s clips. However, since some of the YouTube video links have expired, we can not download the full dataset. Our Kinectic-400 only has 237,644 out of 246,535 training videos and 19,761 validation videos. The training set is about $4\%$ less than that in SlowFast \cite{slowfast} and some of the videos have a duration less than 10s. This leads to an accuracy drop of 0.6\% on Slow-8$\times$8, 1.2\% on SlowFast-4$\times$16 and 0.93\% on X3D-M as we reproduce the results with the officially released codes \cite{slowfastcode}.
\textbf{Charades} \cite{charade} is a multi-label action classification dataset with longer activity duration. The average activity duration is $\sim$30 seconds. The dataset is composed of $\sim$9.8k training videos and 1.8k validation videos in 157 classes. The evaluation metric is mean Average Precision (mAP).
\textbf{AVA} \cite{gu2018ava} is a video dataset for spatio-temporal localization of human actions. It consists of 211k training and 57k validation video segments. We follow previous works \cite{slowfast} to report the mean Average Precision (mAP) on 60 classes using an IoU threshold of 0.5.

\textbf{Implementation Details.}
For single-pathway structures, we adopt Slow 8$\times$8 and X3D-M as our backbone. For multiple-pathway structures we use SlowFast 4$\times$16 due to the limitation of GPU memory. For Slow and SlowFast, the width factor $\gamma_{w}$ is uniformly sampled from $[0.63, 1.0]\times$. The spatial resolution factor is $\gamma_{s} \in \{0.63, 0.80, 1.0\}$ (corresponding to $\{142, 178, 224\}$) and the temporal resolution factor is $\gamma_{t} \in \{0.4, 0.63, 1.0\}$ (corresponding to $\{3, 5, 8\}$). For X3D-M, the width factor $\gamma_{w}$ is uniformly sampled from $[0.63, 1.0]\times$. The spatial resolution factor is $\gamma_{s} \in \{0.63, 0.71, 0.86, 1.0\}$ (corresponding to $\{142, 160, 192, 224\}$) and the temporal resolution factor is $\gamma_{t} \in \{0.4, 0.6, 0.8, 1.0\}$ (corresponding to $\{6, 9, 12, 16\}$). Other training settings are the same as the official codes \cite{slowfastcode}. 

\subsubsection{Main results}
\textbf{Evaluation on Kinetics-400.}
In Fig. \ref{fig:3d-kinetics}, we report the results of MutualNet on different backbones along with its separated trained counterparts. The original results (reported in paper) are denoted as ``-P'' and our reproduced results using official code \cite{slowfastcode} are denoted as ``-R''. We report both results to have a fair comparison since we can not reproduce the original results due to lack of data. For results of MutualNet, we use different line colors to show different dimensions for accuracy-efficiency trade-off. \textcolor{red}{Red} means the spatial resolution is reduced to meet the dynamic resource budget in this range. Similarly, \textcolor{green}{green} stands for temporal resolution and \textcolor{blue}{blue} stands for network width. \textbf{Black} indicates multiple dimensions are involved for one trade-off step.

As shown in Fig. \ref{fig:3d-kinetics}, MutualNet consistently outperforms its separately trained counterparts on three network backbones. It achieves significant improvements over our reproduced results and shows clear advantages over the reported results in the paper. The improvement is even more significant (3.5\% on Slow backbone and 1.6\% on SlowFast backbone) for small resource budgets. This is because our method allows the model to find a better width-spatial-temporal trade-off at each resource budget. And the mutual learning scheme can transfer the knowledge in large configurations to small models to further improve its performance. On X3D-M backbone, our method achieves consistent improvements under different budgets. Note that X3D finds the best-performed model configuration by a search process. It trains many model configurations independently and choose the best one, while MutualNet train all configurations jointly which saves training time and improves the overall performance.

\begin{figure}[t]
    \centering
    \includegraphics[width=\linewidth]{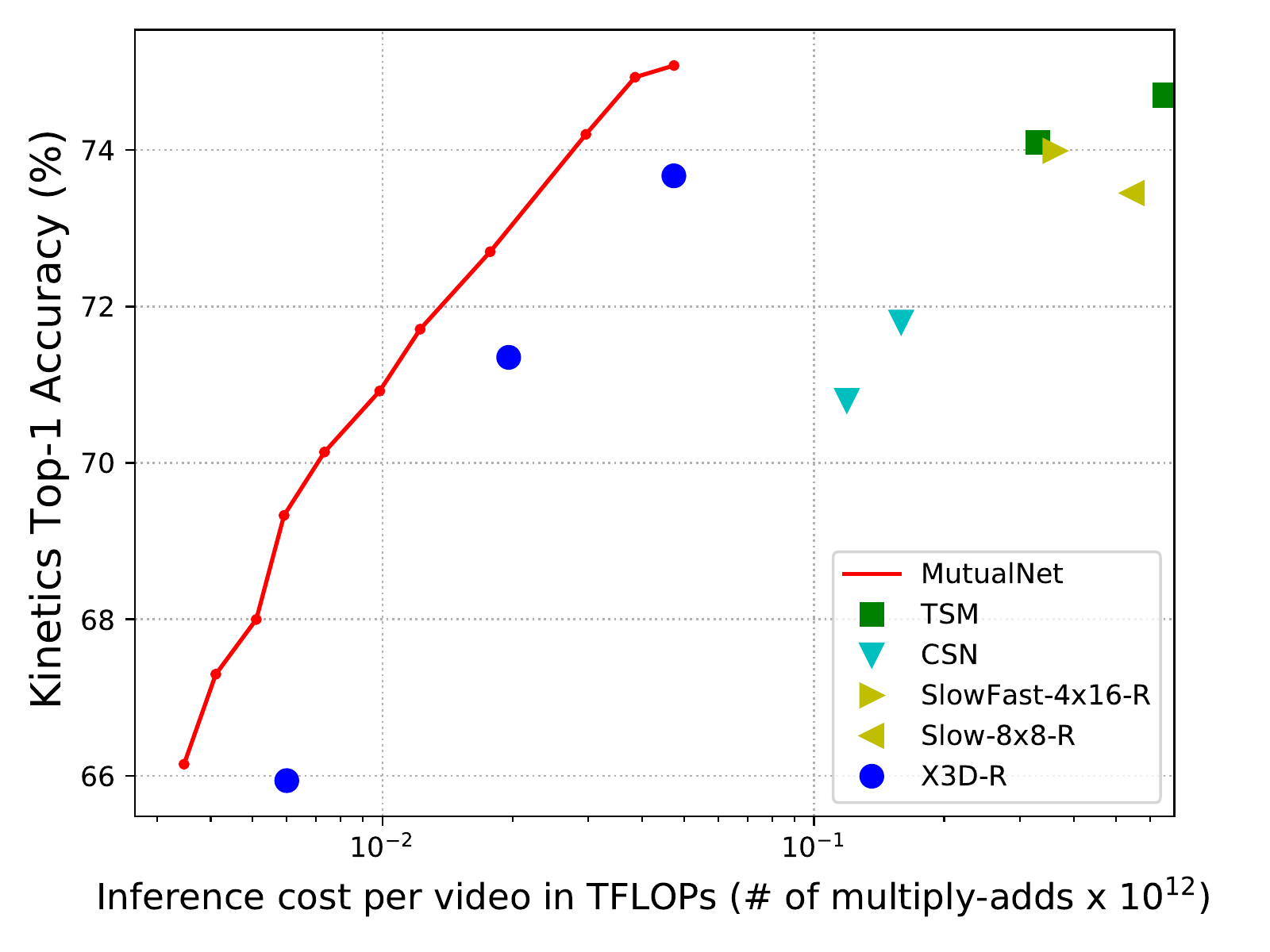}
    \caption{Comparison of MutualNet-X3D-M with state-of-the-art 3D networks. X-axis is in log-scale.}
    \label{fig:3dsota}
\end{figure}

\textbf{Comparison with state-of-the-art.} Based on X3D-M backbone, we compare MutualNet with state-of-the-art methods \cite{CSN, TSM, slowfast, x3d} for action recognition in Fig. \ref{fig:3dsota}. The results are based on 10-view testing. Note that \textbf{the x-axis is in log-scale for better visualization.} We can see that based on the state-of-the-art structure (X3D), MutualNet substantially outperforms previous works (including X3D). This reveals MutualNet is a general training framework and can benefit from improved model structures.

\subsubsection{Transfer learning}
\textbf{Evaluation on Chrades.}
We finetune the models trained on Kinetics-400 on Charades. For SlowFast models, we use the pre-trained models reproduced by us for a fair comparison. For MutualNet models, we do not perform adaptive training during finetuning. That means both SlowFast models and MutualNet models follow the same finetuning process on Charades. The only difference is the pre-trained models. We follow the training settings in the released codes \cite{slowfastcode}. Since we train the model on 4 GPUs, we reduce the batch-size and base learning rate by half following the linear scaling rule \cite{lrscale}. All other settings remain unchanged. As can be seen in Table \ref{tab:charades}, MutualNet model outperforms its counterpart (Slow-8$\times$8) by 0.9\% without increasing the computational cost. Note that the only difference lies in the pre-trained model, so the improvement demonstrate that our method helps the network learn effective and well-generalized representations which are transferable across different datasets.

\begin{table}[!t]
\Large
    \centering
    \caption{Comparison of different models on Charades.}
    \resizebox{\linewidth}{!}{
    \begin{tabular}{l|c|c|c}
    \hline
    model & pretrain & mAP & GFLOPs$\times$views \\
    \hline
    CoViAR, R-50 \cite{coviar} & ImageNet & 21.9 & N/A \\
    Asyn-TF, VGG16 \cite{asyn} & ImageNet & 22.4 & N/A \\
    MultiScale TRN \cite{trn} & ImageNet & 25.2 & N/A \\
    Nonlocal, R-101 \cite{nonlocal} & ImageNet+Kinetics & 37.5 & $544 \times 30$ \\
     \hline
     Slow-8$\times$8 & Kinetics & 34.7 & $54.5 \times 30$ \\
    MutualNet-Slow-8$\times$8 & Kinetics & \textbf{35.6} & \textbf{$54.5 \times 30$} \\
    \hline
    \end{tabular}
    }
    
    \label{tab:charades}
\end{table}

\textbf{Evaluation on AVA Detection.}
Similar to the experiments in Charades, we follow the same training settings as the released SlowFast codes \cite{slowfastcode}. The detector is similar to Faster R-CNN \cite{fasterrcnn} with minimal modifications adopted for video. The region proposals are pre-computed by an off-the-shelf person detector. Experiments are conducted on AVA v2.1. All models are trained on a 4-GPU machine for 20 epochs with a batch-size of 32. The base learning rate is 0.05 with linear warm-up for the first 5 epochs. The learning rate is reduced by a factor of 10 at the 10th and 15th epochs. Both SlowFast pre-trained models and MutualNet pre-trained models are finetuned following the standard training procedure; the only difference is the pre-trained models. As shown in Table \ref{tab:ava}, MutualNet pre-trained model also outperforms SlowFast and previous methods. Note that only the pre-trained weights are different in the experiments, so the improvements are not marginal and clearly demonstrate the effectiveness of the learned representations.
\begin{table}[!t]
 \small
    \centering
    \caption{Comparison of different models on AVA v2.1.}
    \begin{tabular}{l|c|c|c}
    \hline
    model & flow & pretrain & mAP \\ 
    \hline
    I3D \cite{i3d} &  & Kinetics-400 & 14.5 \\
    I3D \cite{i3d} & \checkmark & Kinetics-400 & 15.6 \\
    ACRN, S3D \cite{actorcentric} & \checkmark & Kinetics-400 & 17.4 \\
    ATR, R-50+NL \cite{atr} & & Kinetics-400 & 20.0 \\
     \hline
     Slow-8$\times$8 & & Kinetics-400 & 20.2 \\
    MutualNet-Slow-8$\times$8 & & Kinetics-400 & \textbf{20.6} \\ 
    \hline
    \end{tabular}
    \label{tab:ava}
\end{table}

\section{Conclusion}\label{sec:conclude}
This paper presents a method to mutually learn from different model configurations. After training, the model can do inference at different resource budgets to achieve adaptive accuracy-efficiency trade-offs. Extensive experiments have shown that it significantly improves inference performance per FLOP on various network structures, datasets and tasks. The mutual learning scheme is also a promising training strategy for boosting single network performance. The generality of the proposed method allows it to translate well to generic problem domains.


%

\ifCLASSOPTIONcaptionsoff
  \newpage
\fi




\bibliographystyle{IEEEtran}
{\bibliography{ref}}
\end{document}